\documentclass[preprint,3p,times,twocolumn]{elsarticle}

\usepackage{amsthm,array}
\usepackage{graphicx,amssymb}
\usepackage{latexsym}
\usepackage[ruled,vlined,linesnumbered]{algorithm2e}
\usepackage{amsmath}
\usepackage[all]{xy}
\usepackage{multirow}
\usepackage{mathnotation}
\usepackage{url}
\usepackage{color}
\usepackage[nomargin,inline,index]{fixme}
\fxusetheme{color}

\newcommand{\tax}{\mathit{ax}}

\newcommand{\mo}{\mathcal{O}}
\newcommand{\mt}{\mathcal{T}}
\newcommand{\ma}{\mathcal{A}}

\newcommand{\md}{\mathcal{D}}
\newcommand{\mb}{\mathcal{B}}
\newcommand{\Te}{P}
\newcommand{\Tne}{N}
\newcommand{\te}{p}
\newcommand{\tne}{n}
\newcommand{\qry}{Q}
\newcommand{\ent}{E}

\newcommand{\mD}{{\bf{D}}}
\newcommand{\dx}{{\bf{D^{P}}}}
\newcommand{\dnx}{{\bf{D^{N}}}}
\newcommand{\dz}{{\bf{D^\emptyset}}}
\newcommand{\dxi}[1]{{\bf{D^{P}_{#1}}}}
\newcommand{\dnxi}[1]{{\bf{D^{N}_{#1}}}}
\newcommand{\dzi}[1]{{\bf{D^\emptyset_{#1}}}}

\newcommand{\mT}{\mathcal{T}}
\newcommand{\mA}{\mathcal{A}}
\newcommand{\mCN}{\mathcal{CN}}
\newcommand{\mRN}{\mathcal{RN}}
\newcommand{\mIN}{\mathcal{IN}}
\newcommand{\mI}{\mathcal{I}}

\newtheorem{definition}{Definition}
\newtheorem{corollary}{Corollary}
\newtheorem{property}{Property}
\newtheorem{proposition}{Proposition}
\newtheorem{example}{Example}

\journal{Web Semantics: Science, Services and Agents on the World Wide Web}

\begin{document}

\begin{frontmatter}


\title{Interactive ontology debugging: two query strategies for efficient fault localization\tnoteref{ext}} 


\author[aau]{Kostyantyn Shchekotykhin\corref{C}}
\ead{kostya@ifit.uni-klu.ac.at}
\author[aau]{Gerhard Friedrich}
\ead{gerhard@ifit.uni-klu.ac.at}
\author[aau]{Philipp Fleiss\fnref{grant}}
\ead{pfleiss@ifit.uni-klu.ac.at}
\author[aau]{Patrick Rodler\fnref{grant}}
\ead{prodler@ifit.uni-klu.ac.at}


\fntext[grant]{The research project is funded by grants of the Austrian Science Fund (Project V-Know, contract 19996)} 
\tnotetext[ext]{This article is a substantial extension of the preliminary results published in \emph{Proceedings of the 9th International Semantic Web Conference (ISWC 2010)}~\cite{ShchekotykhinF10}.}
\cortext[C]{Corresponding author at: Alpen-Adria Universit\"at, Universit\"atsstrasse 65-67, 9020 Klagenfurt, Austria. Tel: +43 463 2700 3768, Fax:++43 463 2700 993768}


\address[aau]{Alpen-Adria Universit\"at, Universit\"atsstrasse 65-67, 9020 Klagenfurt, Austria}

\begin{abstract}
Effective debugging of ontologies is an important prerequisite for their broad application, especially in areas that rely on everyday users to create and maintain knowledge bases, such as the Semantic Web. In such systems ontologies capture formalized vocabularies of terms shared by its users. However in many cases users have different local views of the domain, i.e. of the context in which a given term is used. Inappropriate usage of terms together with natural complications when formulating and understanding logical descriptions may result in faulty ontologies.
Recent ontology debugging approaches use diagnosis methods to identify causes of the faults. In most debugging scenarios these methods return many alternative diagnoses, thus placing the burden of fault localization on the user. This paper demonstrates how the target diagnosis can be identified by performing a sequence of observations, that is, by querying an oracle about entailments of the target ontology. To identify the best query we propose two query selection strategies: a simple ``split-in-half" strategy and an entropy-based strategy. The latter allows knowledge about typical user errors to be exploited to minimize the number of queries. Our evaluation showed that the entropy-based method significantly reduces the number of required queries compared to the ``split-in-half" approach. We experimented with different probability distributions of user errors and different qualities of the a-priori probabilities. Our measurements demonstrated the superiority of entropy-based query selection even in cases where all fault probabilities are equal, i.e.\ where no information about typical user errors is available.
\end{abstract}


\end{frontmatter}

\section{Introduction}
Ontology acquisition and maintenance are important prerequisites for the successful application of semantic systems in areas such as the Semantic Web. 
However, as state of the art ontology extraction methods cannot automatically acquire ontologies in a complete and error-free fashion, users of such systems must formulate and correct logical descriptions on their own. In most of the cases these users are domain experts who have little or no experience in expressing knowledge in representation languages like OWL~2~DL~\cite{Grau2008a}.
Studies in cognitive psychology, e.g.~\cite{Ceraso71,Johnson1999}, indicate that humans make systematic errors while formulating or interpreting logical descriptions, with the results presented in~\cite{Rector2004,Roussey2009} confirming that these observations also apply to ontology development. 
Moreover, the problem gets even more if an ontology is developed by a group of users, such as OBO Foundry\footnote{\url{http://www.obofoundry.org}} or NCI Thesaurus\footnote{\url{http://ncit.nci.nih.gov}}, is based on a set of imported third-party ontologies, etc. In this case inconsistencies might appear if some user does not understand or accept the \emph{context} in which shared ontological descriptions are used. Therefore, identification of erroneous ontological definitions is a difficult and time-consuming task.

Several ontology debugging methods ~\cite{schlobach2007,Kalyanpur.Just.ISWC07,friedrich2005gdm,Horridge2008} were proposed to simplify ontology development and maintenance. Usually the main aim of debugging is to obtain a consistent and, optionally, coherent ontology. These basic requirements can be extended with additional ones, such as test cases~\cite{friedrich2005gdm}, which must be fulfilled by the \emph{target ontology} $\mo_t$. Any ontology that does not fulfill the requirements is \emph{faulty} regardless of how it was created. For instance, an ontology might be created by an expert 
specializing descriptions of the imported ontologies (top-down) or 
by an inductive learning algorithm from a set of examples (bottom-up).

Note that
even if all requirements are completely specified, 
many logically equivalent target ontologies might exist. 
They may differ in aspects such as the complexity of consistency checks, size or readability. However, selecting between logically equivalent theories based on such measures is out of the scope of this paper. 
Furthermore, although target ontologies may evolve as requirements change over time, we assume that the target ontology remains stable throughout a debugging session. 

Given an set of requirements (e.g.\ formulated by a user) and a faulty ontology, the task of an ontology debugger is to identify the set of alternative diagnoses, where each diagnosis corresponds to a set of possibly faulty axioms. More concretely, a \emph{diagnosis} $\md$ is a subset of an ontology $\mo$ such that one should remove (change) all the axioms of a diagnosis from the ontology (i.e.\ $\mo \setminus \md$) in order to formulate an ontology $\mo'$ that fulfills all the given requirements. 
Only if the set of requirements is complete the only possible ontology $\mo'$ corresponds to the target ontology $\mo_t$. In the following we refer to the removal of a diagnosis from the ontology as a \emph{trivial application} of a diagnosis.
Moreover, in practical applications it might be inefficient to consider all possible diagnoses. Therefore, modern ontology debugging approaches focus on the computation of minimal diagnoses. 
A set of axioms $\md_i$ is a \emph{minimal diagnosis} iff there is no proper subset $\md'_i \subset \md_i$ which is a diagnosis. 
Thus, minimal diagnoses constitute minimal required changes to the ontology.

Application of diagnosis methods can be problematic in the cases for which many alternative minimal diagnoses exist for a given set of test cases and requirements. A sample study of real-world incoherent ontologies, which were used in~\cite{Kalyanpur.Just.ISWC07}, showed that hundreds or even thousands of minimal diagnoses may exist. In the case of the Transportation ontology the diagnosis method was able to identify 1782 minimal diagnoses \footnote{In Section \ref{sect:eval}, we will give a detailed characterization of these ontologies.}. In such situations a simple visualization of all alternative sets of modifications to the ontology is ineffective. 
Thus an efficient debugging method should be able to discriminate between the diagnoses in order to select the \emph{target diagnosis} $\md_t$. Trivial application of $\md_t$ to the ontology $\mo$ allows a user to extend $(\mo \setminus \md_t)$ with a set of additional axioms $EX$ and, thus, to formulate the target ontology $\mo_t$, i.e.\ $\mo_t = (\mo \setminus \md_t) \cup EX$.

One possible solution to the diagnosis discrimination problem would be to order the set of diagnoses by various preference criteria. For instance, Kalyanpur et al.~\cite{Kalyanpur2006} suggest a measure to rank the axioms of a diagnosis depending on their structure, usage in test cases, provenance, and impact in terms of entailments. Only the top ranking diagnoses are then presented to the user.
Of course this set of diagnoses will contain the target diagnosis only in cases where the faulty ontology, the given requirements and test cases provide sufficient data to the appropriate heuristic. However, it is difficult to identify which information, e.g. test cases, is really required to identify the target diagnosis. That is, a user does not know a priori which and how many tests should be provided to the debugger to ensure that it will return the target diagnosis.

In this paper we present an approach for the acquisition of additional information by \emph{generating} a sequence of queries, the answers of which can be used to reduce the set of diagnoses and ultimately identify the target diagnosis. These queries should be answered by an oracle such as a user or an information extraction system. In order to construct queries we exploit the property that different ontologies resulting from trivial applications of different diagnoses entail unequal sets of axioms. 
Consequently, we can differentiate between diagnoses by asking the oracle if the target ontology should entail a set of logical sentences or not. These entailed logical sentences can be generated by the classification and realization services provided in description logic reasoning systems~\cite{Sirin2007Pellet,Haarslev2001,Motik2009}. In particular, the classification process computes a subsumption hierarchy (sometimes also called ``inheritance hierarchy'' of parents and children) for each concept description mentioned in a TBox. For each individual mentioned in an ABox, the realization computes all the concept names of which the individual is an instance~\cite{Sirin2007Pellet}. 

We propose two methods for selecting the next query of the set of possible queries:
The first method employs a greedy approach that selects queries which try to cut the number of diagnoses in half. The second method exploits the fact that some diagnoses are more likely than others because of typical user errors~\cite{Rector2004,Roussey2009}. Beliefs for an error to occur in a given part of a knowledge base, represented as a probability, can be used to estimate the change in entropy of the set of diagnoses if a particular query is answered. In our evaluation the fault probabilities of axioms are estimated by the type and number of the logical operators employed. For example, roughly speaking, the greater the number of logical operators and the more complex these operators are, the greater the fault probability of an axiom. For assigning prior fault probabilities to diagnoses we employ the fault probabilities of axioms. Of course other methods for guessing prior fault probabilities, e.g. based on context of concept descriptions, measures suggested in the previous work~\cite{Kalyanpur2006}, etc., can be easily integrated in our framework. 
Given a set of diagnoses and their probabilities the method selects a query which minimizes the expected entropy of a set of diagnoses after an oracle answers a query, i.e.\ maximizes the information gain. An oracle should answer such queries until a diagnosis is identified whose probability is significantly higher than those of all other diagnoses. This diagnosis is most likely to be the target diagnosis. 

In 
the first evaluation scenario 
we compare the performance of both methods in terms of the number of queries needed to identify the target diagnosis.
The evaluation is performed using generated examples as well as real-world ontologies presented in Tables~\ref{tab:motivation} and~\ref{tab:bigstats}. In the first case we alter a consistent and coherent ontology with additional axioms to generate conflicts that result in a predefined number of diagnoses of a required length. Each faulty ontology is then analyzed by the debugging algorithm using entropy, greedy and ``random'' strategies, where the latter selects queries at random. The evaluation results show that in some cases the entropy-based approach is almost $60\%$ better than the greedy one whereas both approaches clearly outperformed the random strategy.

In the second evaluation scenario we investigate the robustness of the entropy-based strategy with respect to variations in the prior fault probabilities. We analyze the performance of entropy-based and greedy strategies on real-world ontologies by simulating different types of prior fault probability distributions as well as the ``quality" of these probabilities that might occur in practice. In particular, we identify the cases where all prior fault probabilities are (1) equal, 
(2) ``moderately'' varied or (3) ``extremely'' varied. 
Regarding the ``quality" of the probabilities we investigate cases where the guesses based on the prior diagnosis probabilities are good, average or bad. 
The results show that the entropy method outperforms ``split-in-half" in almost all of the cases, namely when the target diagnosis is located in the more likely two thirds of the minimal diagnoses. 
In some situations the entropy-based approach achieves even twice the performance of the greedy one. 
Only in cases where the initial guess of the prior probabilities is very vague (the bad case), \emph{and} the number of queries needed to identify the target diagnosis is low, ``split-in-half" may save on average one query. 
However, if the number of queries increases, the performance of the entropy-based query selection increases compared to the ``split-in-half" strategy. We observed that if the number of queries is greater than 10, the entropy-based method is preferable even if the initial guess of the prior probabilities is bad. This is due to the effect that the initial bad guesses are improved by the Bayes-update of the diagnoses probabilities as well as an ability of the entropy-based method to stop in the cases when a probability of some diagnosis is above an acceptance threshold predefined by the user. 
Consequently, entropy-based query selection is robust enough to handle different prior fault probability distributions. 

Additional experiments performed on big real-world ontologies demonstrate the scalability of the suggested approach. In our experiments we were able to identify the target diagnosis in an ontology with over 33000 axioms using entropy-based query selection in only 190 seconds using an average of five queries.

The remainder of the paper is organized as follows: Section~\ref{sect:example} presents two introductory examples as well as the basic concepts. The details of the entropy-based query selection method are given in Section~\ref{sect:theory}. Section~\ref{sect:impl} describes the implementation of the approach and is followed by evaluation results in Section~\ref{sect:eval}. The paper concludes with an overview of related work.

\section{Motivating examples and basic concepts}\label{sect:example}

We begin by presenting the fundamentals of ontology diagnosis and then show how queries and answers can be generated and employed to differentiate between sets of diagnoses. 

\subsection{Description logics}

Since the underlying knowledge representation method of ontologies in the Semantic Web is based on description logics, we start by briefly introducing the main concepts, employing the usual definitions as in \cite{borgida96,bad03}. A knowledge base is comprised of two components, namely a TBox (denoted by $\mT$) and a ABox ($\mA$). The TBox defines the terminology whereas the ABox contains assertions about named individuals in terms of the vocabulary defined in the TBox. The vocabulary consists of concepts, denoting sets of individuals, and roles, denoting binary relationships between individuals. These concepts and roles may be either atomic or complex, the latter being obtained by employing description operators. The language of descriptions is defined recursively by starting from a schema $S = (\mCN, \mRN, \mIN)$ of disjoint sets of names for concepts, roles, and individuals. Typical operators for the construction of complex descriptions are $C \sqcup D$ (disjunction), $C \sqcap D$ (conjunction), $\neg C$ (negation), $\forall R.C$ (concept value restriction), and $\exists R.C$(concept exists restriction), where $C$ and $D$ are elements of $\mCN$ and $R \in \mRN$.

Knowledge bases are defined by a finite set of logical sentences. Sentences regarding the TBox are called terminological axioms whereas sentences regarding the ABox are called assertional axioms. Terminological axioms are expressed by $C \sqsubseteq D$ (Generalized Concept Inclusion) which corresponds to the logical implication. Let $a,b \in \mIN$ be individual names. $C(a)$ and $R(a,b)$ are thus assertional axioms.

Concepts (rsp.\ roles) can be regarded as unary (rsp.\ binary) predicates. Roughly speaking description logics can be seen as fragments of first-order predicate logic (without considering transitive closure or special fixpoint semantics). These fragments are specifically designed to ensure decidability or favorable computational costs.

The semantics of description terms are usually given using an interpretation $\mI = \langle \Delta^\mI, (\cdot)^\mI \rangle$, where $\Delta^\mI$ is a domain (non-empty universe) of values, and $(\cdot)^\mI$ is a function that maps every concept description to a subset of $\Delta^\mI$, and every role name to a subset of $\Delta^\mI \times \Delta^\mI$. The mapping also associates a value in $\Delta^\mI$ with every individual name in $\mIN$.
An interpretation $\mI$ is a model of a knowledge base iff it satisfies all terminological axioms and assertional axioms. A knowledge base is satisfiable iff a model exists.
A concept description $C$ is coherent (satisfiable) w.r.t.\ a TBox $\mT$, if  a model $\mI$ of $\mT$ exists such that $C^\mI \neq \emptyset$. A
TBox is incoherent iff an incoherent concept description exists.

\subsection{Diagnosis of ontologies}

\begin{example} \label{ex:simple} Consider a simple ontology $\mo$ with the  terminology $\mt$:
\begin{center}
\begin{tabular}{cc}
$\tax_1 : A \sqsubseteq B$ & 
$\tax_2 : B \sqsubseteq C$  \\
$\tax_3 : C \sqsubseteq D$ &
$\tax_4 : D \sqsubseteq R$ 
\end{tabular}
\end{center}
and assertions $\ma :\{A(w), \lnot R(w), A(v)\}$. 
\end{example}
\noindent Assume that the user explicitly states that the three assertional axioms should be considered as correct, i.e. these axioms are added to a background theory $\mb$. The introduction of a background theory ensures that the diagnosis method focuses purely on the potentially faulty axioms.

Furthermore, assume that the user requires the currently inconsistent ontology $\mo$ to be consistent. The only irreducible set of axioms (minimal conflict set) that preserves the inconsistency is $CS:\left<\tax_1,\tax_2,\tax_3,\tax_4\right>$. That is, one has to modify or remove the axioms of at least one of the following diagnoses
\begin{equation*}
\md_1:\left[\tax_1\right]\quad \md_2:\left[\tax_2\right]\quad \md_3:\left[\tax_3\right]\quad \md_4:\left[\tax_4\right]
\end{equation*}
to restore the consistency of the ontology. However, it is unclear which of the ontologies $\mo_i = \mo \setminus \md_i$ obtained by application of diagnoses from the set ${\bf D}: \{\md_1,\dots, \md_4\}$ is the target one.

\begin{definition}\label{def:targetontology}

A target ontology $\mo_t$ is a set of logical sentences characterized by a set of background axioms $\mb$, a set of sets of logical sentences $\Te$ that must be entailed by $\mo_t$ and the set of sets of logical sentences $\Tne$ that must not be entailed by $\mo_t$. 

A target ontology $\mo_t$ must fulfill the following necessary requirements:
\begin{itemize}
	\item $\mo_t$ must be satisfiable (optionally coherent)
	\item $\mb \subseteq \mo_t$
	\item $\mo_t \models \te\quad \forall \te \in \Te$
	\item $\mo_t \not\models \tne\quad \forall \tne \in \Tne$  
\end{itemize}
\end{definition}
Given $\mb,$ $\Te,$ and $\Tne$, an ontology $\mo$ is faulty iff $\mo$ does not fulfill all the necessary requirements of the target ontology.

Note that the approach presented in this paper can be used with any knowledge representation language for which there exists a sound and complete procedure to decide whether $\mo\models\tax$ and the entailment operator $\models$ is extensive, monotone and idempotent. For instance, these requirements are fulfilled by all subsets of OWL 2 which are interpreted under OWL Direct Semantics.

Definition~\ref{def:targetontology} allows a user to identify the target diagnosis $\md_t$ by providing sufficient information about the target ontology in the sets $\mb, \Te$ and $\Tne$. For instance, if in Example~\ref{ex:simple} the user provides the information that $\mo_t \models \setof{B(w)}$ and $\mo_t \not\models \setof{C(w)}$, the debugger will return only one diagnosis, namely $\md_2$. Application of this diagnosis results in a consistent ontology $\mo_2 = \mo\setminus\md_2$ that entails $\setof{B(w)}$ because of $\tax_1$ and the assertion $A(w)$. In addition, $\mo_2$ does not entail $\setof{C(w)}$ since $\mo_2 \cup \{\neg C(w)\}$ is consistent and, moreover, $ \{ \lnot R(w), \tax_4, \tax_3 \} \models \setof{\lnot C(w)}$. All other ontologies $\mo_i = (\mo\setminus\md_i)$ obtained by the application of the diagnoses $\md_1,\md_3$ and $\md_4$ do not fulfill the given requirements, since $\mo_1 \cup \setof{B(w)}$ is inconsistent and therefore any consistent extension of $\mo_1$ cannot entail $\setof{B(w)}$. As both $\mo_3$ and $\mo_4$ entail $\setof{C(w)}$, $\mo_2$ corresponds to the target diagnosis $\mo_t$.

\begin{definition}\label{def:diag}
Let $\tuple{\mo, \mb,\Te,\Tne}$ be a diagnosis problem instance, where $\mo$ is an ontology, $\mb$ a background theory, $\Te$ a set of sets of logical sentences which must be entailed by the target ontology $\mo_t$, and $\Tne$ a set of sets of logical sentences which must \emph{not} be entailed by $\mo_t$.

A set of axioms $\md\subseteq\mo$ is a diagnosis iff the set of axioms $\mo \setminus \md$ can be extended by a logical description $EX$ such that:
\begin{enumerate}
    \item $(\mo \setminus \md) \cup \mb \cup EX$ is consistent (and coherent if required)
	\item $(\mo \setminus \md) \cup \mb \cup EX \models \te\quad \forall\te \in \Te$
    \item $(\mo \setminus \md) \cup \mb \cup EX \not\models \tne\quad \forall\tne \in \Tne$
\end{enumerate}
\end{definition}

A diagnosis $\md_i$ defines a partition of the ontology $\mo$ where  each axiom $\tax_j \in \md_i$ is a candidate for changes by the user and each axiom $\tax_k \in \mo\setminus\md_i$ is correct. 
If $\md_t$ is the set of axioms of $\mo$ to be changed (i.e.\ $\md_t$ is the target diagnosis) then the target ontology $\mo_t$ is $(\mo \setminus \md_t) \cup \mb \cup EX$ for some $EX$ defined by the user. 

In the following we assume the background theory $\mb$ together with the sets of logical sentences in the sets $\Te$ and $\Tne$ always allow formulation of the target ontology. Moreover, a diagnosis exists iff a target ontology exists.

\begin{proposition}
A diagnosis $\md$ for a diagnosis problem instance $\tuple{\mo,\mb,\Te,\Tne}$ exists iff 

\begin{equation*}
\mb \cup \bigcup_{\te \in \Te}\te
\end{equation*}
is consistent (coherent) and 
\begin{equation*}
\forall \tne \in \Tne \;:\;  \mb \cup \bigcup_{\te \in \Te}\te \not\models \tne
\end{equation*}
\end{proposition}

The set of all diagnoses is complete in the sense that at least one diagnosis exists where the ontology resulting from the trivial application of a diagnosis is a subset of the target ontology:

\begin{proposition}
Let $\mD \neq \emptyset$ be the set of all diagnoses for a diagnosis problem instance $\tuple{\mo,\mb,\Te,\Tne}$ and $\mo_t$ the target ontology. Then a diagnosis $\md_t \in \mD$ exists s.t.\ $(\mo \setminus \md_t) \subseteq \mo_t$. 
\end{proposition}

The set of all diagnoses can be characterized by the set of minimal diagnoses. 

\begin{definition}
A diagnosis $\md$ for a diagnosis problem instance $\tuple{\mo, \mb,\Te,\Tne}$ is a \emph{minimal diagnosis} iff there is no $\md^\prime \subset \md$ such that $\md^\prime$ is a diagnosis. 
\end{definition}


\begin{proposition}
Let $\tuple{\mo, \mb,\Te,\Tne}$ be a diagnosis problem instance. For every diagnosis $\md$ there is a minimal diagnosis $\md'$ s.t. $\md' \subseteq \md$. 
\end{proposition}

\begin{definition}
A diagnosis $\md$ for a diagnosis problem instance $\tuple{\mo, \mb,\Te,\Tne}$ is a \emph{minimum cardinality diagnosis} iff there is no diagnosis $\md^\prime$ such that $|\md^\prime|<|\md|$. 
\end{definition}

To summarize, a diagnosis describes which axioms are candidates for modification. Despite the fact that multiple diagnoses may exist, some are more preferable than others. E.g.\ minimal diagnoses require minimal changes, i.e.\ axioms are not considered for modification unless there is a reason. Minimal cardinality diagnoses require changing a minimal number of axioms. The actual type of error contained in an axiom is irrelevant as the concept of diagnosis defined here does not make any assumptions about errors themselves. There can, however, be instances where an ontology is faulty and the empty diagnosis is the only minimal diagnosis, e.g. if some axioms are missing and nothing must be changed. 

The extension $EX$ plays an important role in the ontology repair process, suggesting axioms that should be added to the ontology. For instance, in Example~\ref{ex:simple} the user requires that the target ontology \emph{must not} entail $\setof{B(w)}$ but has to entail $\setof{B(v)}$, that is $\Tne=\{\{B(w)\}\}$ and $\Te=\{\{B(v)\}\}$. Because, the example ontology $\mo$ is inconsistent some sentences must be changed. The consistent ontology $\mo_1=\mo \setminus \md_1$, neither entails $\setof{B(v)}$ nor $\setof{B(w)}$ (in particular  $\mo_1 \models \setof{\lnot B(w)}$). Consequently, $\mo_1$ has to be extended with a set $EX$ of logical sentences in order to entail $\setof{B(v)}$. This set of logical sentences can be approximated with $EX=\{B(v)\}$. $\mo_1 \cup EX$ is satisfiable, entails $\setof{B(v)}$ but does not entail $\setof{B(w)}$. 
All other ontologies $O_i = \mo \setminus \md_i, \; i=2,3,4$ are consistent but entail $\setof{B(w), B(v)}$ and must be rejected because of the monotonic semantics of description logic. That is, there is no such extension $EX$ that $(\mo_i \cup EX) \not\models \setof{B(w)}$. Therefore, the diagnosis $\md_1$ is the minimum cardinality diagnosis which allows the formulation of the target ontology. 
Note that formulation of the complete extension is impossible, since our diagnosis approach deals with changes to existing axioms and does not learn new axioms.

The following corollary characterizes diagnoses without employing the true extension $EX$ to formulate the target ontology. The idea is to use the sentences which must be entailed by the target ontology to approximate $EX$ as shown above.
\begin{corollary}
Given a diagnosis problem instance $\tuple{\mo, \mb,\Te,\Tne}$, a set of axioms $\md \subseteq \mo$ is a diagnosis iff 
\begin{flalign*}
(\mo \setminus \md) \cup \mb \cup \bigcup_{\te \in \Te}\te \qquad & \text{(Condition~1)} \\
\text{is satisfiable (coherent) and}\qquad\qquad & \\
\forall \tne \in \Tne\;:\; (\mo \setminus \md) \cup \mb \cup \bigcup_{\te \in \Te}\te \not\models \tne \qquad& \text{(Condition~2)}
\end{flalign*}
 \end{corollary}

\noindent \textbf{Proof sketch:} $(\Rightarrow)$ Let $\md \subseteq \mo$ be a diagnosis for $\tuple{\mo, \mb,\Te,\Tne}$. Since there is an $EX$ s.t.\ $(\mo \setminus \md) \cup \mb \cup EX$ is satisfiable (coherent) and $(\mo \setminus \md) \cup \mb \cup EX \models \te$ for all $\te \in \Te$, it follows that $(\mo \setminus \md) \cup \mb \cup EX \cup \bigcup_{\te \in \Te}\te$ is satisfiable (coherent) and therefore $(\mo \setminus \md) \cup \mb \cup \bigcup_{\te \in \Te}\te$ is satisfiable (coherent). Consequently, the first condition of the corollary is fulfilled. Since $(\mo \setminus \md) \cup \mb \cup EX \models \te$ for all $\te \in \Te$ and $(\mo \setminus \md) \cup \mb \cup EX \not\models \tne$ for all $\tne \in \Tne$ it follows that $(\mo \setminus \md) \cup \mb \cup EX \cup \bigcup_{\te \in \Te}\te \not\models \tne$ for all $\tne \in \Tne$. Consequently, $(\mo \setminus \md) \cup \mb \cup \bigcup_{\te \in \Te}\te \not\models \tne$ for all $\tne \in \Tne$ and the second condition of the corollary is fulfilled. 

$(\Leftarrow)$ Let $\md \subseteq \mo$ and $\tuple{\mo, \mb,\Te,\Tne}$ be a diagnosis problem instance. Without limiting generality let $EX = \Te$. By Condition 1 of the corollary $(\mo \setminus \md) \cup \mb \cup \bigcup_{\te \in \Te}\te$ is satisfiable (coherent). Therefore, for $EX = \Te$ the sentences $(\mo \setminus \md) \cup \mb \cup EX$ are satisfiable (coherent), i.e.\ the first condition for a diagnosis is fulfilled and these sentences entail $\te$ for all $\te \in \Te$ which corresponds to the second condition a diagnosis must fulfill. Furthermore, by Condition 2 of the corollary $(\mo \setminus \md) \cup \mb \cup EX \not\models \tne$ for all $\tne \in \Tne$ holds and therefore the third condition for a diagnosis is fulfilled. Consequently, $\md \subseteq \mo$ is a diagnosis for $\tuple{\mo, \mb,\Te,\Tne}$. $\Box$

\emph{Conflict sets}, which are the parts of the ontology that preserve the inconsistency/incoherency, are usually employed to constrain the search space during computation of diagnoses.
\begin{definition}
Given a diagnosis problem instance $\tuple{\mo,\mb,\Te,\Tne}$,  a set of axioms $CS \subseteq \mo$ is a conflict set iff  $CS \cup \mb \cup \bigcup_{\te \in \Te}\te$ is inconsistent (incoherent) or $\tne \in \Tne$ exists s.t.\ $CS \cup \mb \cup \bigcup_{\te \in \Te}\te \models \tne$.
\end{definition}

\begin{definition}
A conflict set $CS$ for an instance $\tuple{\mo,\mb,\Te,\Tne}$ is minimal iff there is no $CS^\prime \subset CS$ such that $CS^\prime$ is a conflict set.
\end{definition}
A set of minimal conflict sets can be used to compute the set of minimal diagnoses as shown in~\cite{Reiter87}. The idea is that each diagnosis must include at least one element of each minimal conflict set. 
\begin{proposition}\label{prop:hittingset}
$\md$ is a minimal diagnosis for the diagnosis problem instance $\tuple{\mo, \mb,\Te,\Tne}$ iff $\md$ is a minimal hitting set for the set of all minimal conflict sets of $\tuple{\mo, \mb,\Te,\Tne}$.
\end{proposition}
Given a set of sets $\overline S$, a set $H$ is a hitting set of $\overline S$  iff $H \cap S_i \neq \emptyset$ for all $S_i \in \overline S$ and $H \subseteq \bigcup_{S_i \in \overline S} S_i$. 
Most modern ontology diagnosis methods~\cite{schlobach2007,Kalyanpur.Just.ISWC07,friedrich2005gdm,Horridge2008} are implemented according to Proposition~\ref{prop:hittingset} and differ only in details, such as how and when (minimal) conflict sets are computed, the order in which hitting sets are generated, etc.

\subsection{Differentiating between diagnoses}\label{sect:discrimination}

The diagnosis method usually generates a set of diagnoses for a given diagnosis problem instance. Thus, in Example~\ref{ex:simple} an ontology debugger returns a set of four minimal diagnoses $\{\md_1, \dots, \md_4\}$. 
As explained in the previous section, additional information, i.e.\ sets of sets of logical sentences $\Te$ and $\Tne$, can be used by the debugger to reduce the set of diagnoses. However, in the general case the user does not know which sets $\Te$ and $\Tne$ to provide to the debugger such that the target diagnosis will be identified. Therefore, the debugger should be able to identify sets of logical sentences on its own and only ask the user or some other oracle, whether these sentences \emph{must} or \emph{must not} be entailed by the target ontology. To generate these sentences the debugger can apply each of the diagnoses in $\mD=\{\md_1,\dots,\md_n\} $ and obtain a set of ontologies $\mo_i = \mo \setminus \md_i\:,\: i=1,\dots, n$ that fulfill the user requirements. For each ontology $\mo_i$ a description logic reasoner can generate a set of entailments such as entailed subsumptions provided by the classification service and sets of class assertions provided by the realization service. 
These entailments can be used to discriminate between the diagnoses, as different ontologies entail different sets of sentences due to extensivity of the entailment relation. 
Note that in the examples provided in this section we consider only two types of entailments, namely subsumption and class assertion.
In general, the approach presented in this paper is not limited to these types and can use all of the entailment types supported by a reasoner. 
\begin{table}%
\centering
\begin{tabular}{cl}
Ontology & Entailments \\ \hline
$\mo_1$&$\emptyset$ \\
$\mo_2$&$\{B(w)\}$ \\
$\mo_3$&$\{B(w),C(w)\}$ \\
$\mo_4$&$\{B(w),C(w), D(w)\}$ \\\hline
\end{tabular}
\caption{Entailments of ontologies $\mo_i = (\mo \setminus \md_i)\:, \: i=1,\dots, 4$  in Example~\ref{ex:simple} returned by realization.}
\label{tab:entailex1}
\end{table}

For instance, in Example~\ref{ex:simple} for each ontology $\mo_i = (\mo \setminus \md_i) \:,\: i=1\dots 4$ the realization service of a reasoner returns the set of class assertions presented in Table~\ref{tab:entailex1}.
Without any additional information the debugger cannot decide which of these sentences must be entailed by the target ontology. To obtain this information the diagnosis method must query an oracle that can specify whether the target ontology entails some set of sentences or not. E.g.\ the debugger could ask an oracle if $\setof{D(w)}$ is entailed by the target ontology ($\mo_t \models \setof{D(w)}$). 
If the answer is \emph{yes}, then $\setof{D(w)}$ is added to $\Te$ and $\md_4$ is considered as the target diagnosis. All other diagnoses are rejected because $(\mo \setminus \md_i) \cup \mb \cup \{D(w)\}$ for $i=1,2,3$ is inconsistent.  If the answer is \emph{no}, then $\setof{D(w)}$ is added to $\Tne$ and $\md_4$ is rejected as $(\mo \setminus \md_4) \cup \mb \models \setof{D(w)}$ and we have to ask the oracle another question. In the following we consider a query $\qry$ as a set of logical sentences such that $\mo_t \models \qry$ holds iff $\mo_t \models q_i$ for all $q_i \in \qry$.

\begin{property}
Given a diagnosis problem instance $\tuple{\mo, \mb,\Te,\Tne}$, a set of diagnoses $\mD$, a set of logical sentences $\qry$
representing the query $(\mo_t \models \qry)\,$ and an oracle able to evaluate the query:

If the oracle  answers \emph{yes} then every diagnosis $\md_i \in \mD$ is a diagnosis for $\Te \cup \{ \qry \}$ iff both conditions hold:
\begin{align*}
(\mo \setminus \md_i)& \cup \mb \cup \bigcup_{\te \in \Te}\te \cup \qry \; \emph{is consistent (coherent)}\\
\forall \tne \in \Tne & \;:\;(\mo \setminus \md_i)  \cup \mb \cup \bigcup_{\te \in \Te}\te \cup \qry \not\models \tne
\end{align*}

If the oracle answers \emph{no} then every diagnosis $\md_i \in \mD$ is a diagnosis for $\Tne \cup \{ \qry \}$ iff both conditions hold:
\begin{align*}
(\mo \setminus \md_i)& \cup \mb \cup \bigcup_{\te \in \Te}\te\; \emph{is consistent (coherent)} \\
\forall \tne \in 
(\Tne \cup \{ \qry \}) & \;:\; (\mo \setminus \md_i) \cup \mb \cup \bigcup_{\te \in \Te}\te \not\models \tne
\end{align*}
\end{property}

In particular, a query partitions the set of diagnoses $\mD$ into three disjoint subsets.

\begin{definition}\label{def:partition}
For a query $\qry$, each diagnosis $\md_i \in \mD$ of a diagnosis problem instance $\tuple{\mo, \mb,\Te,\Tne}$ can be assigned to one of the three sets $\dx$, $\dnx$ or $\dz$
where
\begin{itemize}
\item $\md_i \in \dx$ iff it holds that
\begin{equation*}
(\mo \setminus \md_i) \cup \mb \cup \bigcup_{\te \in \Te}\te \models \qry
\end{equation*}
\item $\md_i \in \dnx$ iff it holds that
\begin{equation*}
(\mo \setminus \md_i) \cup \mb \cup \bigcup_{\te \in \Te}\te \cup  \qry
\end{equation*}
is inconsistent (incoherent). 
\item $\md_i \in \dz$ iff $\md_i \in \mD\setminus\left(\dx \cup \dnx\right)$
\end{itemize}
\end{definition}

Given a diagnosis problem instance we say that the diagnoses in $\dx$ predict a positive answer (\emph{yes}) as a result of the query $\qry$, diagnoses in $\dnx$ predict a negative answer (\emph{no}), and diagnoses in $\dz$ do not make any predictions. 

\begin{property}\label{prop:removeDiag}
Given a diagnosis problem instance $\tuple{\mo, \mb,\Te,\Tne}$, a set of diagnoses $\mD$, a query $\qry$ and an oracle:

If the oracle answers \emph{yes} then the set of rejected diagnoses is $\dnx$ and the set of remaining diagnoses is $\dx\cup \dz$.

If the oracle answers \emph{no} then the set of rejected diagnoses is $\dx$ and the set of remaining diagnoses is $\dnx \cup \dz$.
\end{property}

Consequently, given a query $\qry$ either $\dx$ or $\dnx$ is eliminated but $\dz$ always remains after the query is answered. For generating queries we have to investigate for which subsets $\dx, \dnx \subseteq \mD$ a query exists that can differentiate between these sets. A straight forward approach is to investigate all possible subsets of $\mD$. In our evaluation we show that this is feasible if we limit the number $n$ of minimal diagnoses to be considered during query generation and selection. E.g.\ for $n=9$, the algorithm has to verify $512$ possible partitions in the worst case.

Given a set of diagnoses $\mD$ for the ontology $\mo$, a set $\Te$ of sets of sentences that must be entailed by the target ontology $\mo_t$ and a set of background axioms $\mb$, the set of partitions $\bf{PR}$ for which a query exists can be computed as follows:
\begin{enumerate}
	\item Generate the power set $\pset{\mD}$, $\bf{PR}\leftarrow\emptyset$
    \item Assign an element of $\pset{\mD}$ to the set $\dxi{i}$ and generate a set of common entailments $\ent_{i}$ of all ontologies $(\mo \setminus \md_j) \cup \mb \cup \bigcup_{\te \in \Te}\te$, where $\md_j \in \dxi{i}$
    \item If $\ent_{i} = \emptyset$, then reject the current element $\dxi{i}$, 
    i.e.\ set $\pset{\mD} \leftarrow \pset{\mD} \setminus \{\dxi{i}\}$ and goto Step 2. Otherwise set $\qry_i \leftarrow \ent_i$.
    \item Use Definition~\ref{def:partition} and the query $\qry_i$ to classify the diagnoses $\md_k \in \mD \setminus \dxi{i}$ into the sets $\dxi{i}$, $\dnxi{i}$ and $\dzi{i}$. The generated partition is added to the set of partitions $\mathbf{PR} \leftarrow \mathbf{PR} \cup \{\tuple{\qry_i,\dxi{i},\dnxi{i},\dzi{i}}\}$ and set $\pset{\mD} \leftarrow \pset{\mD}\setminus \{\dxi{i}\}$. If $\pset{\mD} \neq \emptyset$ then go to Step 2.
\end{enumerate}

In Example~\ref{ex:simple} the set of diagnoses $\mD$ of the ontology $\mo$ contains 4 elements. Therefore, the power set $\pset{\mD}$ includes 15 elements $\setof{\{\md_1\},\setof{\md_2},\dots, \setof{\md_1,\md_2,\md_3,\md_4}}$, assuming we omit the element corresponding to $\emptyset$ as it does not contain any diagnoses to be evaluated. Moreover, assume that $\Te$ and $\Tne$ are empty. 
In each iteration an element of $\pset{\mD}$ is assigned to the set $\dxi{i}$. For instance, the algorithm assigns $\dxi{1}=\{\md_1,\md_2\}$. In this case the set of common entailments is empty as $(\mo\setminus\md_1) \cup \mb$ has no entailed 
sentences 
(see Table~\ref{tab:entailex1}). Therefore, the set $\{\md_1,\md_2\}$ is rejected and removed from $\pset{\mD}$. Assume that in the next iteration the algorithm selects $\dxi{2}=\{\md_2,\md_3\}$. 
In this case the set of common entailments $\ent_2 = \setof{B(w)}$ is not empty and so $\qry_2=\{B(w)\}$. 
The remaining diagnoses $\md_1$ and $\md_4$ are classified according to Definition~\ref{def:partition}. That is, the algorithm selects the first diagnosis $\md_1$ and verifies whether $(\mo \setminus \md_1) \cup \mb \models \setof{B(w)}$. 
Given the negative answer of the reasoner, the algorithm checks if $(\mo \setminus \md_1) \cup \mb \cup \{B(w)\}$ is inconsistent. Since the condition is satisfied the diagnosis $\md_1$ is added to the set $\dnxi{2}$. 
The second diagnosis $\md_4$ is added to the set $\dxi{2}$ as it satisfies the first requirement $(\mo \setminus \md_4) \cup \mb \models \{B(w)\}$. 
The resulting partition $\tuple{\{B(w)\},\{\md_2,\md_3,\md_4\},\{\md_1\}, \emptyset}$ is added to the set $\bf{PR}$. 

However, a query need not include all of the entailed sentences. If a query $\qry$ partitions the set of diagnoses into $\dx$, $\dnx$ and $\dz$ and an (irreducible) subset $\qry' \subset \qry$ exists which preserves the partition then it is sufficient to query $\qry'$. In our example, $\qry_2:\{B(w),C(w)\}$ can be reduced to its subset $\qry'_2:\{C(w)\}$.  If there are multiple irreducible subsets that preserve the partition then we select one of them.

All of the queries and their corresponding partitions generated in Example~\ref{ex:simple} are presented in Table~\ref{tab:example1}. 
Given these queries the debugger has to decide which one should be asked first in order to minimize the number of queries to be answered. A popular query selection heuristic (called ``split-in-half'') prefers queries which allow half of the diagnoses to be removed from the set $\mD$ regardless of the answer of an oracle. 

Using the data presented in Table~\ref{tab:example1}, the ``split-in-half'' heuristic determines that asking the oracle if $(\mo_t \models \{C(w)\})$ is the best query (i.e.\ the reduced query $\qry_2$), as two diagnoses from the set $\mD$ are removed regardless of the answer. Assuming that $\md_1$ is the target diagnosis, then an oracle will answer $no$ to our question (i.e.\ $\mo_t \not\models \setof{C(w)}$). Based on this feedback, the diagnoses $\md_3$ and $\md_4$ are removed according to Property~\ref{prop:removeDiag}. Given the updated set of diagnoses $\mD$ and $\Te=\setof{\setof{C(w)}}$ the partitioning algorithm returns the only partition $\tuple{\setof{B(w)},\setof{\md_2},\setof{\md_1}, \emptyset}$. The heuristic then selects the query $\setof{B(w)}$, which is also answered with $no$ by the oracle. 
Consequently, $\md_1$ is identified as the only remaining minimal diagnosis. 

\begin{table*}[tb]
\centering
\begin{tabular}{@{\extracolsep{8pt}} llll}
Query & $\dx$  & $\dnx$ &  $\dz$ \\\hline
$\qry_1:\{B(w)\}$ & $\{\md_2,\md_3,\md_4\}$ & $\{\md_1\}$ & $\emptyset$ \\
$\qry_2:\{B(w),C(w)\}$ & $\{\md_3, \md_4 \}$   & $\{\md_1, \md_2\}$ & $\emptyset$ \\
$\qry_3:\{B(w),C(w),Q(w)\}$ & $\{\md_4\}$         & $\{\md_1, \md_2, \md_3 \}$ &
$\emptyset$ \\\hline
\end{tabular}
\caption{Possible queries in Example~\ref{ex:simple}}\label{tab:example1}
\end{table*}

In general, if $n$ is the number of diagnoses and we can split the set of diagnoses in half with each query, then the minimum number of queries is
$log_2{n}$. Note that this minimum number of queries can only be achieved when all minimal diagnoses are considered at once, which is intractable even for relatively small values of $n$. 

However, in case probabilities of diagnoses are known we
can reduce the number of queries by utilizing two effects: 
\begin{enumerate}
	\item We can exploit diagnoses probabilities to assess the likelihood of each answer and the expected value of the information contained in the set of diagnoses after an answer is given. 
    \item Even if multiple diagnoses remain, further query generation may not be required if one diagnosis is highly probable and all other remaining diagnoses are highly improbable.
\end{enumerate}

\begin{example}\label{ex:complex} Consider an ontology $\mo$ with the terminology $\mt$:
\begin{center}
\begin{tabular}{ll}
$\tax_1 : A_1 \sqsubseteq A_2 \sqcap M_1 \sqcap M_2$ & 
$\tax_4 : M_2 \sqsubseteq \forall s.A \sqcap D$ \\
$\tax_2 : A_2 \sqsubseteq \lnot\exists s.M_3 \sqcap \exists s.M_2$ & 
$\tax_5 : M_3 \equiv B \sqcup C$ \\
$\tax_3 : M_1 \sqsubseteq \lnot A \sqcap B$ & \\
\end{tabular}
\end{center}
and the background theory containing the assertions $\ma :\{A_1(w), A_1(u), s(u,w)\}$. 
\end{example}
The ontology is inconsistent and 
the set of minimal conflict sets $CS =\{\left<\tax_1,\tax_3,\tax_4\right>, $ $\left<\tax_1,\tax_2,\tax_3,\tax_5\right>\}$.
To restore consistency, the user should modify all axioms of at least one minimal diagnosis: 
\begin{align*}
\md_1&:\left[\tax_1\right]  &\md_3&:\left[\tax_4,\tax_5\right]\\
\md_2&:\left[\tax_3\right]  &\md_4&:\left[\tax_4,\tax_2\right]
\end{align*}

\begin{figure}[b]
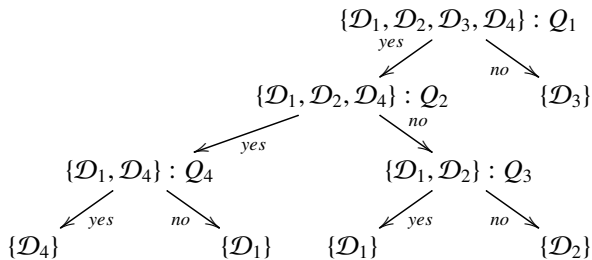

\centering
\xygraph{
!{<0cm,0cm>;<1.4cm,0cm>:<0cm,1cm>::}
!{(0,0) }*+{\{\md_4\}}="d4"
!{(2,0) }*+{\{\md_1\}}="d14"
!{(3,0) }*+{\{\md_1\}}="d13"
!{(5,0) }*+{\{\md_2\}}="d2"
!{(1,1) }*+{\{\md_1, \md_4\} : \qry_4}="x4"
!{(4,1) }*+{\{\md_1, \md_2\} : \qry_3}="x3"
!{(3,2) }*+{\{\md_1, \md_2, \md_4\} : \qry_2}="x2"
!{(5,2) }*+{\{\md_3\}}="d3"
!{(4,3)}*+{\{\md_1, \md_2, \md_3, \md_4\} : \qry_1}="x1"
"x4":"d4"^{yes}
"x4":"d14"_{no}
"x3":"d13"^{yes}
"x3":"d2"_{no}
"x2":"x4"^{yes}
"x2":"x3"^{no}
"x1":"d3"_{no}
"x1":"x2"_{yes}
}
\caption{The search tree of the greedy algorithm} \label{fig:greedy:ex}
\end{figure}

Following the same approach as in the first example, we compute a set of possible queries and corresponding partitions using the algorithm presented above.
A set of possible irreducible queries for Example~\ref{ex:complex} and their partitions are presented in Table~\ref{tab:example2}.
These queries partition the set of diagnoses ${\bf D}$ in a way that makes the application of myopic strategies, such as ``split-in-half'', inefficient.
A greedy algorithm based on such a heuristic would first select the first query $\qry_1$,
since there is no query that cuts the set of diagnoses in half. If $\md_4$ is the target diagnosis then $\qry_1$ will be answered with $yes$ by an oracle (see Figure~\ref{fig:greedy:ex}). In the next iteration the algorithm would also choose a suboptimal query, the first untried query $\qry_2$, since there is no partition that divides the diagnoses $\md_1$, $\md_2$, and $\md_4$ into two groups of equal size. Once again, the oracle answers $yes$, and the algorithm identifies query $\qry_4$ to differentiate between $\md_1$ and $\md_4$.

\begin{table*}[!htb]
\centering
\begin{tabular}{@{\extracolsep{8pt}}llll}
Query & $\dx$  & $\dnx$ &  $\dz$ \\ \hline
$\qry_1:\{B \sqsubseteq M_3\}$ & $\{\md_1,\md_2,\md_4\}$ & $\{\md_3\}$ & $\emptyset$\\
$\qry_2:\{B(w)\}$ & $\{\md_3, \md_4\}$ & $\{\md_2\}$ & $\{\md_1\}$ \\
$\qry_3:\{M_1 \sqsubseteq B\}$ & $\{\md_1,\md_3,\md_4\}$ & $\{\md_2\}$ & $\emptyset$\\
$\qry_4:\{M_1(w), M_2(u)\}$ & $\{\md_2,\md_3,\md_4\}$ & $\{\md_1\}$ & $\emptyset$  \\
$\qry_5:\{A(w)\}$ & $\{\md_2\}$ & $\{\md_3, \md_4\}$ & $\{\md_1\}$\\
$\qry_6:\{M_2\sqsubseteq D\}$ & $\{\md_1,\md_2\}$ & $\emptyset$ & $\{\md_3, \md_4\}$\\
$\qry_7:\{M_3(u)\}$ & $\{\md_4\}$ & $\emptyset$ & $\{\md_1, \md_2, \md_3\}$\\\hline
\end{tabular}
\caption{Possible queries in Example 2}\label{tab:example2}
\end{table*}

However, in real-world settings the assumption that all axioms fail with the same probability is rarely the case. For example, Roussey et al.~\cite{Roussey2009} present a list of ``anti-patterns'' where an anti-pattern is a set of axioms, such as $\{C1\sqsubseteq\forall R.C2, C1\sqsubseteq\forall R.C3, C2\equiv\lnot C3\}$ that corresponds to a minimal conflict set. The study performed by~\cite{Roussey2009} shows that such conflict sets often occur in practice due to frequent misuse of certain language constructs like quantification or disjointness. Such studies are ideal sources for estimating prior fault probabilities. However, this is beyond the scope of this paper. 
 
Our approach for computing the prior fault probabilities of axioms is inspired by Rector et al.~\cite{Rector2004} and considers the syntax of a knowledge representation language, such as restrictions, conjunction, negation, etc. 
For instance, if a user frequently changes the universal to the existential quantifier and vice versa in order to restore coherency, then we can assume that axioms including such restrictions are more likely to fail than the other ones. 
In~\cite{Rector2004} the authors report that in most cases inconsistent ontologies are created because users (a) mix up $\forall r.S$ and $\exists r.S$,  (b) mix up $\lnot\exists r.S$ and $\exists r.\lnot S$, (c) mix up $\sqcup$ and $\sqcap$, (d) wrongly assume that classes are disjoint by default or overuse disjointness, or (e) wrongly apply negation. Observing that misuses of quantifiers are more likely than other failure patterns one might find that the axioms $\tax_2$ and $\tax_4$ are more likely to be faulty than $\tax_3$ (because of the use of quantifiers), whereas $\tax_3$ is  more likely to be faulty than $\tax_5$ and $\tax_1$ (because of the use of negation). 

Detailed justifications of diagnoses probabilities are given in the next section. However, let us assume some probability distribution of the faults according to the observations presented above such that: 
(a) the diagnosis $\md_2$ is the most probable one, i.e.\ single fault diagnosis of an axiom containing a negation; 
(b) although $\md_4$ is a double fault diagnosis, it follows $\md_2$ closely as its axioms contain quantifiers; 
(c) $\md_1$ and $\md_3$ are significantly less probable than $\md_4$ because conjunction/disjunction in $\tax_1$ and $\tax_5$ have a significantly lower fault probability than negation in $\tax_3$. 
Taking this information into account asking query $\qry_1$ is essentially useless because it is highly probable that the target diagnosis is either $\md_2$ or $\md_4$ and, therefore, it is highly probable that the oracle will respond with $yes$. 
Instead, asking $\qry_3$ is more informative because regardless of the answer we can exclude one of the highly probable diagnoses, i.e.\ either $\md_2$ or $\md_4$. 
If the oracle responds to $\qry_3$ with $no$ then $\md_2$ is the only remaining diagnosis. 
However, if the oracle responds with $yes$, diagnoses $\md_4$, $\md_3$, and $\md_1$ remain, where $\md_4$ is significantly more probable compared to diagnoses $\md_3$ and $\md_1$. 
If the difference between the probabilities of the diagnoses is high enough such that $\md_4$ can be accepted as the target diagnosis, no additional questions are required. 
Obviously this strategy can lead to a substantial reduction in the number of queries compared to myopic approaches as we demonstrate in our evaluation.

Note that in real-world application scenarios failure patterns and their probabilities can be discovered by analyzing the debugging actions of a user in an ontology editor, like Prot\'eg\'e. Learning of fault probabilities can be used to ``personalize'' the query selection algorithm to prefer user-specific faults. However, as our evaluation shows, even a rough estimate of the probabilities is capable of outperforming the ``split-in-half'' heuristic.

\section{Entropy-based query selection}\label{sect:theory}

To select the best query we exploit a-priori failure probabilities of each axiom derived from the syntax of description logics or some other knowledge representation language, such as OWL. 
That is, the user is able to specify own beliefs in terms of the probability of syntax element such as $\forall$, $\exists$, $\sqcap$, etc. being erroneous; alternatively, the debugger can compute these probabilities by analyzing the frequency of various syntax elements in the target diagnoses of different debugging sessions. If no failure information is available then the debugger can initialize all of the probabilities with some small value. Compared to statistically well-founded probabilities, the latter approach provides a suboptimal but useful diagnosis discrimination process, as discussed in the evaluation.  

Given the failure probabilities of all syntax elements $se \in {\bf S}$ of a knowledge representation language used in $\mo$, we can compute the failure probability of an axiom $tax_i \in \mo$
\begin{equation*}
p(\tax_i) = p(F_{se_1} \cup F_{se_2} \cup \dots \cup F_{se_n})
\end{equation*}
where $F_{se_1} \dots F_{se_n}$ represent the events that the occurrence of a syntax element $se_j$ in $\tax_i$ is faulty. E.g.\ for $ax_2$ of Example 2 $p(ax_2) = p( F_\sqsubseteq \cup F_\lnot \cup F_\exists \cup F_\sqcap \cup F_\exists)$.  Assuming that each occurrence of a syntax element fails independently, i.e.\ an erroneous usage of a syntax element $se_k$ makes it neither more nor less probable that an occurrence of syntax element $se_j$ is faulty, the failure probability of an axiom is computed as:
\begin{equation}
\label{eq:axiom:prob}
p(\tax_i) = 1 - \prod_{se \in {\bf S}} (1-F_{se})^{c(se)}
\end{equation}
where $c(se_j)$ returns number of occurrences of the syntax element $se_j$ in an axiom $\tax_i$.
If among other failure probabilities the user states that $p(F_\sqsubseteq)=0.001, p(F_\lnot)=0.01, p(F_\exists)=0.05$ and $p(F_\sqcap)=0.001$ then $p(\tax_2) = p(F_\sqsubseteq \cup F_\lnot \cup F_\exists \cup F_\sqcap \cup F_\exists) = 0.108$.

Given the failure probabilities $p(\tax_i)$ of axioms, the diagnosis algorithm first calculates the a-priori probability $p(\md_j)$ that $\md_j$ is the target diagnosis. Since all axioms fail independently, this  probability can be computed as~\cite{dekleer1987}:
\begin{equation}
p(\md_j) = \prod_{\tax_n\ \in \md_j}{p(\tax_n)} \prod_{\tax_m\ \in \mo\setminus\md_j}{1-p(\tax_m)}
\label{eq:dprob}
\end{equation}

The prior probabilities for diagnoses are then used to initialize an iterative algorithm that includes two main steps: (a) the selection of the best query and (b) updating the diagnoses probabilities given query feedback.

According to information theory the best query is the one that, given the answer of an oracle, minimizes the expected entropy of the set of diagnoses~\cite{dekleer1987}. Let $p(\qry_i = yes)$ be the probability that query $\qry_i$ is answered with $yes$ and $p(\qry_i = no)$ be the probability for the answer $no$. Furthermore, let $p(\md_j | \qry_i = yes)$ be the probability of diagnosis $\md_j$ after the oracle answers $yes$ and $p(\md_j | \qry_i = no)$ be the probability after the oracle answers $no$. The expected entropy after querying $\qry_i$ is:
\begin{align*}
H_e(\qry_i) =  & \sum_{v \in \setof{yes,no}} p(\qry_i = v) \times \\
  &-\sum_{\md_j \in \mD} p(\md_j | \qry_i = v) \log_2 p(\md_j | \qry_i = v)
\end{align*}

Based on a one-step-look-ahead information theoretic measure, the query which minimizes the expected entropy is considered best. This formula can be simplified to the following score function~\cite{dekleer1987} which we use to evaluate all available queries and select the one with the minimum score to maximize information gain:
\begin{align}
\begin{split}
sc(\qry_i) = \sum_{v \in \setof{yes,no}} \bigl[p(\qry_i=v)\log_2&{p(\qry_i=v)}\bigr] \\
& + p(\dzi{i}) + 1
\end{split}
\label{eq:score}
\end{align}
where $v \in \setof{yes,no}$ is a feedback of an oracle and ${\bf D^{\emptyset}_i}$ is the set of diagnoses which do not make any predictions for the query $\qry_i$. The probability of the set of diagnoses $p(\dzi{i})$ as well as of any other set of diagnoses ${\bf D_{i}}$ like $\dxi{i}$ and $\dnxi{i}$ is computed as: 
\begin{equation*}
p({\bf D_{i}}) = \sum_{\md_j \in {\bf D_{i}}}{p(\md_j)}
\end{equation*}
because by Definition~\ref{def:diag}, each diagnosis uniquely partitions all of the axioms of an ontology $\mo$ into two sets, correct and faulty, and thus all diagnoses are mutually exclusive events. 

Since, for a query $\qry_i$, the set of diagnoses $\mD$ can be partitioned into the sets $\dxi{i}$, $\dnxi{i}$ and $\dzi{i}$, the probability that an oracle will answer a query $\qry_i$ with either $yes$ or $no$ can be computed as:
\begin{align}
\begin{split}
p&(\qry_i=yes)= p(\dxi{i}) + p(\dzi{i})/2 \\
p&(\qry_i=no)= p(\dnxi{i}) + p(\dzi{i})/2
\end{split}
\label{eq:dupdate:prob}
\end{align}

Clearly this assumes that for each diagnosis of $\dzi{i}$ \emph{both outcomes are equally likely} and thus the probability that the set of diagnoses $\dzi{i}$ predicts either $\qry_i=yes$ or $\qry_i=no$ is $p(\dzi{i})/2$.

Following feedback $v$ for a query $\qry_s$, i.e.\ $\qry_s = v$, the probabilities of the diagnoses must be updated to take the new information into account. The update is made using Bayes' rule for each $\md_j \in \mD$:
\begin{equation}
p(\md_j|\qry_s=v) = \frac{p(\qry_s=v|\md_j)p(\md_j)}{p(\qry_s = v)}
\label{eq:dupdate}
\end{equation}
where the denominator $p(\qry_s = v)$ is known from the query selection step (Equation~\ref{eq:dupdate:prob}) and $p(\md_j)$ is either a prior probability (Equation~\ref{eq:dprob}) or is a probability calculated using Equation~\ref{eq:dupdate} after a previous iteration of the debugging algorithm. We assign $p(\qry_s=v|\md_j)$ as follows:
\begin{equation*}
p(\qry_s=v|\md_j) = 
\begin{cases}
1, & \mbox{if $\md_j$ predicted $\qry_s=v$;}\\
0, & \mbox{if $\md_j$ is rejected by $\qry_s=v$;}\\
\frac{1}{2}, & \mbox{if $\md_j \in \dzi{s}$}
\end{cases} 
\end{equation*}

\vspace{5pt}\noindent\textbf{Example 1 (continued)} 
Suppose that the debugger is not provided with any information about possible failures and therefore assumes that all syntax elements fail with the same probability $0.01$ and therefore $p(\tax_i)=0.01$ for all $\tax_i \in \mo$. Using Equation~\ref{eq:dprob} we can calculate probabilities for each diagnosis. 
For instance, $\md_1$ suggests that only one axiom $\tax_1$ should be modified by the user. Hence, we can calculate the probability of diagnosis $D_1$ as  
$p(\md_1) = p(\tax_1)(1-p(\tax_2))(1-p(\tax_3))(1-p(\tax_4)) = 0.0097$.
All other minimal diagnoses have the same probability, since every other minimal diagnosis suggests the modification of one axiom. To simplify the discussion we only consider minimal diagnoses for query selection. Therefore, the prior probabilities of the diagnoses can be normalized to $p(\md_j)=p(\md_j)/\sum_{\md_j \in \mD}{p(\md_j)}$ and are equal to $0.25$.

Given the prior probabilities of the diagnoses and a set of queries (see Table~\ref{tab:example1}) we evaluate the score function (Equation~\ref{eq:score}) for each query. E.g. for the first query $\qry_1:\{B(w)\}$ the probability $p(\dz)=0$ and the probabilities of both the positive and negative outcomes are: 
$p(\qry_1=1)=p(\md_2)+p(\md_3)+p(\md_4) = 0.75$ and $p(\qry_1=0)=p(\md_1) = 0.25$. Therefore the query score is $sc(\qry_1)=0.1887$. 

The scores computed during the initial stage (see Table~\ref{tab:example1:costs1}) suggest that $\qry_2$ is the best query. Taking into account that $\md_1$ is the target diagnosis the oracle answers $no$ to the query. The additional information obtained from the answer is then used to update the probabilities of diagnoses using the Equation~\ref{eq:dupdate}. 
Since $\md_1$ and $\md_2$ predicted this answer, their probabilities are updated, $p(\md_1) = p(\md_2) = 1/p(\qry_2=1)=0.5$. 
The probabilities of diagnoses $\md_3$ and $\md_4$ which are rejected by the oracle's answer are also updated, $p(\md_3) = p(\md_4) = 0$.

In the next iteration the algorithm recomputes the scores using the updated probabilities. The results show that $\qry_1$ is the best query. The other two queries $\qry_2$ and $\qry_3$ are irrelevant since no information will be gained if they are asked. 
Given the oracle's negative feedback to $\qry_1$, we update the probabilities $p(\md_1) = 1$ and $p(\md_2) = 0$. 
In this case the target diagnosis $\md_1$ was identified using the same number of steps as the ``split-in-half'' heuristic. 

However, if the user specifies that the first axiom is more likely to fail, e.g.\ $p(\tax_1) = 0.025$, then $\qry_1:\{B(w)\}$ will be selected first (see Table~\ref{tab:example1:costs2}). The recalculation of the probabilities given the negative outcome $\qry_1=0$ sets $p(\md_1) = 1$ and $p(\md_2)=p(\md_3)=p(\md_4)=0$. Therefore the debugger identifies the target diagnosis in only one step.

\begin{table}[tb]
\centering
\begin{tabular}{@{\extracolsep{8pt}}lcc}
Query & Initial score& $\; \qry_2 = yes \;$  \\ \hline
$\qry_1:\{B(w)\}$ & 0.1887 & \textbf{0}  \\
$\qry_2:\{C(w)\}$ & \textbf{0}      & 1  \\
$\qry_3:\{Q(w)\}$ & 0.1887 & 1  \\ \hline
\end{tabular}
\caption{Expected scores for minimized queries ($p(\tax_i)=0.01$)}\label{tab:example1:costs1}
\end{table}
\begin{table}[tb]
\centering
		\begin{tabular}{@{\extracolsep{8pt}}lc}
Query & Initial score   \\ \hline
$\qry_1:\{B(w)\}$ & \textbf{0.250}   \\
$\qry_2:\{C(w)\}$ & 0.408   \\
$\qry_3:\{Q(w)\}$ & 0.629  \\ \hline
		\end{tabular}
        \caption{Expected scores for minimized queries ($p(\tax_1)=0.025$, $p(\tax_2)=p(\tax_3)=p(\tax_4) = 0.01)$}\label{tab:example1:costs2}
\end{table}

\vspace{5pt}\noindent\textbf{Example 2 (continued)} 
Suppose that in $\tax_4$ the user specified $\forall s.A$ instead of $\exists s.A$ and $\lnot\exists s.M_3$ instead of $\exists s.\lnot M_3$ in $\tax_2$. Therefore $\md_4$ is the target diagnosis. Moreover, assume that the debugger is provided with observations of three types of faults: (1) conjunction/disjunction occurs with probability $p_1 = 0.001$, (2) negation $p_2=0.01$, and (3) restrictions $p_3=0.05$. 
Using Equation~\ref{eq:axiom:prob} we can calculate the probability of the axioms containing an error: $p(\tax_1)=0.0019$, $p(\tax_2)=0.1074$, $p(\tax_3)=0.012$, $p(\tax_4)=0.051$, and $p(\tax_5)=0.001$. These probabilities are exploited to calculate the prior probabilities of the diagnoses (see Table~\ref{tab:example2:diagnoses}) and to initialize the query selection process. To simplify matters we focus on the set of minimal diagnoses.

In the first iteration the algorithm determines that $\qry_3$ is the best query and asks the oracle whether $\mo_t \models \setof{M_1 \sqsubseteq B}$ is true or not (see Table~\ref{tab:example2:costs}). The obtained information is then used to recalculate the probabilities of the diagnoses and to compute the next best subsequent query, i.e.\ $\qry_4$, and so on. The query process stops after the third query, since $\md_4$ is the only diagnosis that has the probability $p(\md_4) > 0$. 

\begin{table*}[tb]
	\centering
    	\begin{tabular}{lcccc}
Answers	&	$\md_1$	&	$\md_2$	&	$\md_3$	&	$\md_4$	\\ \hline
Prior	&	0.0970	&	0.5874	&	0.0026	&	0.3130	\\
$\qry_3=yes$	&	0.2352	&	0	&	0.0063	&	0.7585	\\
$\qry_3=yes$, $\qry_4=yes$	&	0	&	0	&	0.0082	&	0.9918	\\
$\qry_3=yes$, $\qry_4=yes$, $\qry_1=yes \quad$	&	0	&	0	&	0	&	1	\\ \hline
\end{tabular}
\caption{Probabilities of diagnoses after answers}\label{tab:example2:diagnoses}
\end{table*}

\begin{table*}[tb]
\centering
\begin{tabular}{lccc}
Queries	&	Initial	&	$\qry_3 = yes$	&	$\qry_3 = yes$, $\qry_4 = yes$	\\ \hline
$\qry_1:\{B \sqsubseteq M_3\}$	&	0.974	&	0.945	& \textbf{	0.931	} \\
$\qry_2:\{B(w)\}$	&	0.151	&	0.713	&	1	\\
$\qry_3:\{M_1 \sqsubseteq B\}$	& \textbf{	0.022	}&	1	&	1	\\
$\qry_4:\{M_1(w), M_2(u)\}$	&	0.540	& \textbf{	0.213	}&	1	\\
$\qry_5:\{A(w)\}$	&	0.151	&	0.713	&	1	\\
$\qry_6:\{M_2\sqsubseteq D\}$	&	0.686	&	0.805	&	1	\\
$\qry_7:\{M_3(u)\}$	&	0.759	&	0.710	&	0.970	\\\hline
\end{tabular}
 \caption{Expected scores for queries}\label{tab:example2:costs}
\end{table*}

Given the feedback of the oracle $\qry_4=yes$ for the second query, the updated probabilities of the diagnoses show that the target diagnosis has a probability of $p(\md_4) = 0.9918$ whereas $p(\md_3)$ is only $0.0082$. 
In order to reduce the number of queries a user can specify a threshold, e.g. $\sigma=0.95$. 
If the absolute difference in probabilities of two most probable diagnoses is greater than this threshold, the query process stops and returns the most probable diagnosis. 
Therefore, in this example the debugger based on the entropy query selection requires less queries than the ``split-in-half'' heuristic. 
Note that already after the first answer $\qry_3=yes$ the most probable diagnosis $\md_4$ is three times more likely than the second most probable diagnosis $\md_1$. Given such a great difference we could suggest to stop the query process after the first answer if the user would set $\sigma=0.65$.

\section{Implementation details}\label{sect:impl}

The iterative ontology debugger (Algorithm~\ref{algo:general}) takes a faulty  ontology $\mo$ as input. Optionally, a user can provide a set of axioms $\mb$ that are known to be correct as well as a set $\Te$ of axioms that must be entailed by the target ontology and a set $\Tne$ of axioms that must not. If these sets are not given, the corresponding input arguments are initialized with $\emptyset$. Moreover, the algorithm takes a set $FP$ of fault probabilities for axioms $\tax_i \in \mo$, which can be computed as described in Section~\ref{sect:theory} by exploiting knowledge about typical user errors. Alternatively, if no estimates of such probabilities are available, all probability values can be initialized using a small constant. We show the results of such a strategy in our evaluation section. 
The two other arguments $\sigma$ and $n$ are used to improve the performance of the algorithm. $\sigma$ specifies the diagnosis acceptance threshold, i.e. the minimum difference in probabilities between the most likely and second-most likely diagnoses.  The parameter $n$ defines the maximum number of most probable diagnoses that should be considered by the algorithm during each iteration.
A further performance gain in Algorithm~\ref{algo:general} can be achieved if we approximate the set of the $n$ most probable diagnoses with the set of the $n$ most probable \emph{minimal} diagnoses, i.e.\ we neglect non-minimal diagnoses. We call this set of at most $n$ most probable minimal diagnoses the \emph{leading diagnoses}. Note, under the reasonable assumption that the fault probability of each axiom $p(\tax_i)$ is less than $0.5$, for every non-minimal diagnosis $ND$ a minimal diagnosis $\md \subset ND$ exists which from  Equation~\ref{eq:dprob} is more probable than $ND$.
Consequently the query selection algorithm presented here operates on the set of minimal diagnoses instead of all diagnoses (i.e. non-minimal diagnoses are excluded). 
However, the algorithm can be adapted with moderate effort to also consider non-minimal diagnoses.

We use the approach proposed by Friedrich et al.~\cite{friedrich2005gdm} to compute diagnoses and employ the combination of two algorithms, \textsc{QuickXplain}~\cite{junker04} and \textsc{HS-Tree}~\cite{Reiter87}. In a standard implementation the latter is a breadth-first search algorithm that takes an ontology $\mo$, sets $\Te$ and $\Tne$, and the maximum number of most probable minimal diagnoses $n$ as an input. The algorithm generates minimal hitting sets using minimal conflict sets, which are computed on-demand. This is motivated by the fact that in some circumstances a subset of all minimal conflict sets is sufficient for generating a subset of all required minimal diagnoses. 
For instance, in Example~\ref{ex:complex} the user wants to compute only $n=2$ leading minimal diagnoses and a minimal conflict search algorithm returns $CS_1$. In this case \textsc{HS-Tree} identifies two required minimal diagnoses $\md_1$ and $\md_2$ and avoiding the computation of the minimal conflict set $CS_2$.
Of course, in the worst case, when all minimal diagnoses have to be computed the algorithm should compute all minimal conflict sets.
In addition, the \textsc{HS-Tree} generation reuses minimal conflict sets in order to avoid unnecessary computations. Thus, in the real-world scenarios we evaluated (see Table~\ref{tab:motivation}), less than 10 minimal conflict sets were contained in the faulty ontologies having at most 13 elements while the maximal cardinality of minimal diagnoses was observed to be at most 9. Therefore, space limitations were not a problem for the breadth-first generation. However, for scenarios involving diagnoses of greater cardinalities iterative-deepening strategies could be applied. 

In our implementation of \textsc{HS-Tree} we use the uniform-cost search strategy. Given additional information in terms of axiom fault probabilities $FP$, the algorithm expands a leaf node in a search-tree if it is an element of the path corresponding to the maximum probability hitting set of minimal conflict sets computed so far.
The probability of each minimal hitting set can be computed using Equation~\ref{eq:dprob}. Consequently, the algorithm computes a set of diagnoses ordered by their probability starting from the most probable one.
\textsc{HS-Tree} terminates if either the $n$ most probable minimal diagnoses are identified or no further minimal diagnoses can be found. Thus the algorithm computes at most $n$ minimal diagnoses regardless of the number of all minimal diagnoses. 

\textsc{HS-Tree} uses \textsc{QuickXplain} to compute required minimal conflicts. 
This algorithm, given a set of axioms $AX$ and a set of correct axioms $\mb$ returns a minimal conflict set $CS \subseteq AX$, or $\emptyset$ if axioms $AX \cup \mb$ are consistent.  
In the worst case, to compute a minimal conflict \textsc{QuickXplain} performs $2k (\log (s/k) + 1)$ consistency checks, where $k$ is the size of the generated minimal conflict set and $s$ is the number of axioms in the ontology. 
In the best case only $\log (s/k) + 2k$ are performed~\cite{junker04}. 
Importantly, the size of the ontology is contained in the $\log$ function. Therefore, the time needed for consistency checks in our test ontologies remained below $0.2$ seconds, even for real world knowledge bases with thousands of axioms. The maximum time to compute a minimal conflict was observed in the Sweet-JPL ontology and took approx.\ 5 seconds (see Table~\ref{tab:stats}).

In order to take past answers into account the \textsc{HS-Tree} updates the prior probabilities of the diagnoses by evaluating Equation~\ref{eq:dupdate}. All required data is stored in the query history $QH$ as well as in the sets $\Te$ and $\Tne$. When complete, \textsc{HS-Tree} returns a set of tuples of the form $\tuple{\md_i, p(\md_i)}$ where $\md_i$ is contained in the set of the $n$ most probable minimal diagnoses (leading diagnoses) and $p(\md_i)$ is its probability calculated using Equation~\ref{eq:dprob} and Equation~\ref{eq:dupdate}. 

\begin{algorithm}[bt]

\caption{\textsc{ontoDebugging$(\mo, \mb, P, N, FP, n, \sigma)$}} \label{algo:general}
\KwIn{ontology $\mo$, set of background axioms $\mb$, set of sets of logical sentences to be entailed $P$, set of sets of logical sentences not to be entailed $N$, set of fault probabilities for axioms $FP$, maximum number of most probable minimal diagnoses $n$, \newline acceptance threshold $\sigma$}
\KwOut{a diagnosis $\md$} 
\SetKwFunction{HSTree}{\normalfont \textsc{HS-Tree}}
\SetKwFunction{computeDataSet}{\normalfont \textsc{computeDataSet}}
\SetKwFunction{getSize}{\normalfont \textsc{getSize}}
\SetKwFunction{uptateProbablities}{\normalfont \textsc{uptateProbablities}}
\SetKwFunction{computePriors}{\normalfont \textsc{computePriors}}
\SetKwFunction{getAnswer}{\normalfont \textsc{getAnswer}}
\SetKwFunction{getMinimalEntopyScore}{\normalfont \textsc{getMinimalScore}}
\SetKwFunction{selectMeasurement}{\normalfont \textsc{selectQuery}}
\SetKwFunction{getDiag}{\normalfont \textsc{mostProbableDiagnosis}}
\SetKwFunction{thresh}{\normalfont \textsc{belowThreshold}}
\SetKwFunction{exit}{exit loop}
\SetKwFunction{getScore}{\normalfont \textsc{getScore}}
\SetKwFunction{getQuery}{\normalfont \textsc{getQuery}}
\SetKwFunction{isempty}{\normalfont \textsc{isQueryEmpty}}

$DP \leftarrow \emptyset$;
$QH \leftarrow \emptyset$;
$T \leftarrow \tuple{\emptyset,\emptyset,\emptyset,\emptyset}$\;
\While{{$\thresh(DP,\sigma) \land \getScore(T)\neq 1$}} {
      $DP \leftarrow \HSTree (\mo, \mb, \Te, \Tne, FP, QH, n)$\;  
    $T \leftarrow \selectMeasurement(DP, \mo, \mb, \Te)$\;
		$Q  \leftarrow \getQuery(T)$\;
    \lIf {$Q = \emptyset$} \textbf{exit loop}\;
    \lIf {$\getAnswer(\mo_t \models \qry)$}{        
        $\Te \leftarrow \Te \cup \setof{\qry}$\;
    }
    \lElse{
        $\Tne \leftarrow \Tne \cup \setof{\qry}$\;
    }
    $QH  \leftarrow QH  \cup \setof{T}$\; 
} 

\Return $\getDiag(DP)$\;
\end{algorithm}

\begin{algorithm}[bt]
\caption{\textsc{selectQuery$(DP, \mo, \mb, \Te)$}} \label{algo:computeQuery}
\KwIn{set $DP$ of tuples $\tuple{\md_i, p(\md_i)}$, ontology $\mo$ , set of background axioms $\mb$,  set of sets of logical sentences that must be entailed by the target ontology $\Te$}
\KwOut{a tuple $\tuple{\qry,\dx,\dnx,\dz}$} 
\SetKwFunction{getScore}{\normalfont \textsc{getScore}}
\SetKwFunction{generate}{\normalfont \textsc{generate}}
\SetKwFunction{create}{\normalfont \textsc{createQuery}}
\SetKwFunction{diag}{\normalfont \textsc{getDiagnoses}}
\SetKwFunction{pop}{pop}
\SetKwFunction{min}{\normalfont \textsc{minimizeQuery}}
\SetKwFunction{minCard}{\normalfont \textsc{minCardinalityQuery}}
\SetKwBlock{query}{function {\normalfont \texttt{selectQuery} ($DP, \mo, \mb, \Te$)} }{end}
\SetKwBlock{part}{function {\normalfont \textsc{generate} ($\dx, D, \mo, \mb, \Te, DP$)} \newline \hspace*{60pt} returns {\normalfont a tuple $\tuple{\qry,\dx,\dnx,\dz}$}}{end}
 $\mD \leftarrow \diag(DP)$\;
 $T \leftarrow \generate(\emptyset, \mD, \mo, \mb, \Te, DP)$\;
 \Return $\min(T)$\; \vspace{7pt}

\part{
    \If {$D = \emptyset$}{
        $\mD \leftarrow \diag(DP)$\;  
        \Return \create($\dx,\mo,\mb,P, \mD$)\;
    }
         
    $\md \leftarrow$ \pop($D$)\;
    $\emph{left} \leftarrow$ \generate($\dx, D, \mo, \mb, \Te, DP$)\;
    $right \leftarrow$ \generate($\dx \cup \setof{\md}, D, \mo, \mb, \Te, DP$)\;
    
    \lIf {\getScore(left, DP) $<$ \getScore(right, DP)}{                
        \Return $\emph{left}$\;
    } \lElse{
        \Return $right$\;
    }
}
\end{algorithm}

In the query-selection phase Algorithm~\ref{algo:general} calls \textsc{selectQuery} function (Algorithm~\ref{algo:computeQuery}) to generate a tuple $T=\tuple{\qry,\dx,\dnx,\dz}$, where $\qry$ is the minimum score query (Equation~\ref{eq:score}) and $\dx,\dnx$ and $\dz$ the sets of diagnoses constituting the partition. The generation algorithm carries out a depth-first search, removing the top element of the set $D$ and calling itself recursively to generate all possible subsets of the leading diagnoses. The set of leading diagnoses $\mD$ is extracted from the set of tuples $DP$ by the \textsc{getDiagnoses} function.
In each leaf node of the search tree the \textsc{generate} function calls \textsc{createQuery}  creates a query given a set of diagnoses $\dx$ by computing common entailments and partitioning the set of diagnoses $\mD \setminus \dx$, as described in Section~\ref{sect:discrimination}. If a query for the set $\dx$ does not exist (i.e.\ there are no common entailments) or $\dx=\emptyset$ then \textsc{createQuery} returns an empty tuple $T=\tuple{\emptyset,\emptyset,\emptyset,\emptyset}$. 
In all inner nodes of the tree the algorithm selects a tuple that corresponds to a query with the minimum score as found using the \textsc{getScore} function. This function may implement the entropy-based measure (Equation~\ref{eq:score}), ``split-in-half'' or any other preference criteria. Given an empty tuple $T=\tuple{\emptyset,\emptyset,\emptyset,\emptyset}$ the function returns the highest possible score of a used measure. 
In general, \textsc{createQuery} is called $2^n$ times, where we set $n=9$ in our evaluation. Furthermore, for each leading diagnosis not in $\dx$, \textsc{createQuery} has to check if the associated query is entailed. If a query is not entailed, a consistency check has to be performed. Entailments are determined by classification/realization and a subset check of the generated sentences. Common entailments are computed by exploiting the intersection of entailments for each diagnosis contained in $\dx$. Note that the entailments for each leading diagnosis are computed just once and reused in for subsequent calls of \textsc{createQuery}. 

In the function \textsc{minimizeQuery}, the query $\qry$ of the resulting tuple $\tuple{\qry,\dx,\dnx,\dz}$ is iteratively reduced by applying \textsc{QuickXplain} such that sets $\dx$, $\dnx$ and $\dz$ are preserved.  
This is implemented by replacing the consistency checks performed by \textsc{QuickXplain} with checks that ensure that the reduction of the query preserves the partition. In order to check if a partition is preserved, a consistency/entailment check is performed for each element in $\dnx$ and $\dz$. Elements of $\dx$ need not be checked because these elements entail the query and therefore any reduction. In the worst case $n (2k \log (s/k) + 2k)$ consistency checks have to be performed in \textsc{minimizeQuery} where $k$ is the length of the minimized query. Entailments of leading diagnoses are reused. 

Algorithm~\ref{algo:general} invokes the function \textsc{getQuery} to obtain the query from the tuple stored in $T$ and calls \textsc{getAnswer} to query the oracle. 
Depending on the answer, Algorithm~\ref{algo:general} extends either the set $\Te$ or the set $\Tne$ and thus excludes diagnoses not compliant with the query answer from the results of \textsc{HS-Tree} in further iterations. Note, the algorithm can be easily adapted to allow the oracle to reject a query if the answer is unknown. In this case the algorithm proceeds with the next best query (w.r.t. the \textsc{getScore} function) until no further queries are available. 

Algorithm~\ref{algo:general} stops if the difference in the probabilities of the top two diagnoses is greater than the acceptance threshold $\sigma$ or if no query can be used to differentiate between the remaining diagnoses (i.e.\ the score of the minimum score query equals to the maximum score of the used measure).
The most probable diagnosis is then returned to the user. If it is impossible to differentiate between a number of highly probable minimal diagnoses, the algorithm returns a set that includes all of them. Moreover, in the first case (termination due to $\sigma$), the algorithm can continue if the user is not satisfied with the returned diagnosis and at least one further query exists. 

Additional performance improvements can be achieved by using greedy strategies in Algorithm~\ref{algo:computeQuery}. The idea is to guide the search such that a leaf node of the left-most branch of a search tree contains a set of diagnoses $\dx$ that might result in a tuple $\tuple{\qry,\dx,\dnx,\dz}$ with a low-score query. This method is based on the property of Equation~\ref{eq:score} that $sc(\qry)=0$ if 
\begin{align*}
\sum_{\md_i \in \dx}p(\md_i)=\sum_{\md_j \in \dnx}p(\md_j)=0.5\quad \textnormal{ and } \quad p(\dz)=0
\end{align*} 
Consequently, the query selection problem can be presented as a two-way number partitioning problem: given a set of numbers, divide them into two sets such that the difference between the sums of the numbers in each set is as small as possible. The Complete Karmarkar-Karp (CKK) algorithm~\cite{Korf1998}, which is one of the best algorithms developed for the two-way partitioning problem, corresponds to an extension of the Algorithm~\ref{algo:computeQuery} with a set differencing heuristic~\cite{kk1986}. The algorithm stops if the optimal solution to the two-way partitioning problem is found or if there are no further subsets to be investigated. In the latter case the best found solution is returned. 

The main drawback of applying CKK to the query selection process is that none of the pruning techniques can be used. Also even if the algorithm finds an optimal solution to the two-way partitioning problem there just might be no query for a found set of diagnoses $\dx$. 
Moreover, since the algorithm is complete it still has to investigate all subsets of the set of diagnoses in order to find the minimum score query. 
To avoid this exhaustive search we extended CKK with an additional termination criterion: the search stops if a query is found with a score below some predefined threshold $\gamma$. In our evaluation section we demonstrate substantial savings by applying the CKK partitioning algorithm. 

To sum up, the proposed method depends on the efficiency of the classification/realization system and consistency/coherency checks given a particular ontology. The number of calls to a reasoning system can be reduced by decreasing the number of leading diagnoses $n$. However, the more leading diagnoses provide the more data for generating the next best query. Consequently, by varying the number of leading diagnoses it is possible to balance runtime with the number of queries needed to isolate the target diagnosis.\footnote{The source code as well as precompiled binaries can be downloaded from \url{http://rmbd.googlecode.com}. The package also includes a Prot\'{e}g\'{e}-plugin implementing the methods as described.} 
\section{Evaluation}\label{sect:eval}
 
\begin{table*}[tb]
 \centering
      \begin{tabular}{@{\extracolsep{-2pt}} llcccccl} 
 &  Ontology  &     DL &      Axioms &  \#C/\#P/\#I   &  \#CS/min/max  & \#D/min/max    &  Domain \\ \hline
1.  &  Chemical   &  $\mathcal{ALCHF}^{(D)}$  & 144     &  48/20/0 &  6/5/6 & 6/1/3 &  Chemical elements     \\ 
2.  &   Koala    &  $\mathcal{ALCON}^{(D)}$  &    44 &   21/5/6  &   3/4/4   &   10/1/3  &   Training \\ 
3.  &  Sweet-JPL   &  $\mathcal{ALCHOF}^{(D)}$ &    2579   &  1537/121/50   &  1/13/13 &  13/1/1   &  Earthscience  \\ 
4.  &  miniTambis     &  $\mathcal{ALCN}$     &    173     &  183/44/0   &  3/2/6 & 48/3/3   &  Biological science  \\ 
5.  &  University   &  $\mathcal{SOIN}^{(D)}$   &   49   &  30/12/4 &  4/3/5 & 90/3/4   &  Training \\  
6.  &  Economy   &  $\mathcal{ALCH}^{(D)}$   &   1781   &  339/53/482     &  8/3/4 & 864/4/8 &  Mid-level   \\ 
7.  &  Transportation&    $\mathcal{ALCH}^{(D)}$   &     1300   &  445/93/183     &  9/2/6 & 1782/6/9   &  Mid-level  \\ \hline
      \end{tabular}
 \caption{Diagnosis results for several of the real-world ontologies presented in~\cite{Kalyanpur.Just.ISWC07}. \#C/\#P/\#I are the number of concepts, properties and individuals in each ontology. \#CS/min/max are the number of conflict sets, and their minimum and maximum cardinality. The same notation is used for diagnoses \#D/min/max. The ontologies are available upon request.} 
 \label{tab:motivation}
\end{table*}

\begin{table*}[tb] 
\begin{tabular}{@{\extracolsep{-1pt}}llccc||ccc} 
   & \multicolumn{1}{c}{}&   \multicolumn{3}{c}{Leading diagnoses}  &  \multicolumn{3}{c}{All diagnoses} \\
Ontology  &  \multicolumn{2}{r}{Consistency}   &  Conflicts    &  Diagnoses &  Consistency    &  Conflicts    &  Diagnoses   \\  \hline
Chemical   &  time   &  0/3/8 &  90/107/128     &  1/97/326   &  0/3/18   & 105/130/179   &  2/126/402  \\  
 &  calls    &  264     &  6   &  8   &  262     &  6   &  7   \\ \cline{2-8}
 &  \multicolumn{3}{c}{runtime: 723}   &   &  \multicolumn{2}{r}{runtime: 892}  &   \\  \hline
Koala &  time   &  0/1/3 &  19/25/30   &  0/11/70 &  0/2/4 &  24/30/37 &  0/12/105   \\  
 &  calls    &  74   &  3   &  10   &  75   &  3   &  11   \\ \cline{2-8}
 &  \multicolumn{3}{c}{runtime: 120}   &   &  \multicolumn{2}{r}{runtime: 148}  &   \\  \hline
Sweet-JPL   &  time   &  1/31/112   &  5185/5185/5185 &  0/586/5332     & 31/106/195     &  5192/5192/5192 &  1/438/5319     \\  
 &  calls    &  187     &  1   &  10   &  195     &  1   &  14   \\ \cline{2-8}
 &  \multicolumn{3}{c}{runtime: 5991}    &   &  \multicolumn{2}{r}{runtime: 6312}      &   \\  \hline
miniTambis     &  time   &  0/5/14   &  84/157/210     &  0/57/504   &  1/5/15   & 88/167/225     &  3/19/537   \\  
 &  calls    &  111     &  3   &  10   &  189     &  3   &  49   \\ \cline{2-8}
 &  \multicolumn{3}{c}{runtime: 586}   &   &  \multicolumn{2}{r}{runtime: 1027}      &   \\  \hline
University   &  time   &  0/2/3 &  31/41/54   &  0/20/157   &  0/2/5 & 37/46/60   &  2/5/200 \\  
 &  calls    &  126     &  4   &  10   &  283     &  4   &  91   \\ \cline{2-8}
 &  \multicolumn{3}{c}{runtime: 205}   &   &  \multicolumn{2}{r}{runtime: 536}  &   \\  \hline
Economy  &  time   &  1/12/26 &  410/460/569   &  0/282/2085     &  1/9/80   & 418/510/681   &  16/25/1929     \\  
 &  calls    &  239     &  6   &  10   &  2064   &  8   &  865     \\ \cline{2-8}
 &  \multicolumn{3}{c}{runtime: 2857}    &   &  \multicolumn{2}{r}{runtime: 25369}    &   \\  \hline
Transportaton   &  time   &  0/11/58 &  237/438/683   &  0/352/3176     &  1/9/130 &  222/429/636   &  16/29/6394     \\  
 &  calls    &  337     &  7   &  10   &  3966   &  9   &  1783   \\ \cline{2-8}
 &  \multicolumn{3}{c}{runtime: 3671}    &   &  \multicolumn{2}{r}{runtime: 65010}    &   \\  \hline
\end{tabular}
 \caption{Min/avg/max time and calls required to compute the nine leading most probable diagnoses as well as all diagnoses for the real-world ontologies. Values are given for each stage, i.e. consistency checking, computation of minimal conflicts and minimal diagnoses, together with the total runtime needed to compute the diagnoses. All time values are 15 trial averages and are given in milliseconds.}
 \label{tab:stats}
\end{table*}

\begin{figure*}[tb]
 \centering
 	\includegraphics[width=10cm]{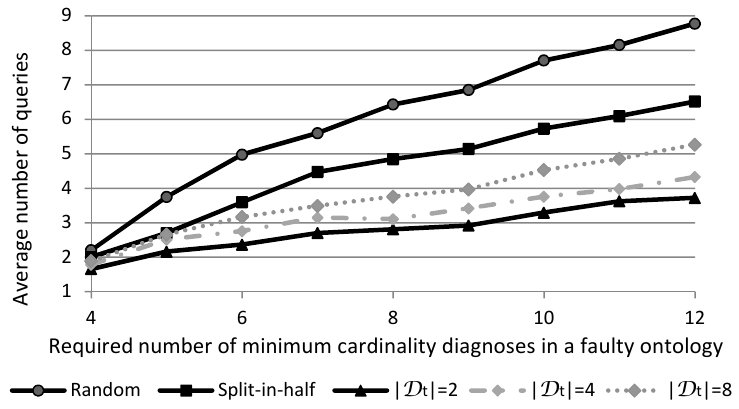}
 \caption{Average number of queries required to select the target diagnosis $\md_t$ with threshold $\sigma=0.95$. Random and ``split-in-half'' are shown for the cardinality of minimal diagnoses $|\md_t|=2$.}
 \label{fig:results}
\end{figure*}

We evaluated our approach using the real-world ontologies presented in Table~\ref{tab:motivation} with the aim of demonstrating its applicability real-world settings. In addition, we employed generated examples to perform controlled experiments where the number of minimal diagnoses and their cardinality could be varied to make the identification of the target diagnosis more difficult. Finally, we carried out a set of tests using randomly modified large real-world ontologies to provide some insights on the scalability of the suggested debugging method.

For the first test we created a generator which takes a consistent and coherent ontology, a set of fault patterns together with their probabilities, the minimum number of minimum cardinality diagnoses $m$, and the required cardinality $|\md_t|$ of these minimum cardinality diagnoses as inputs. We also assumed that the target diagnosis has cardinality $|\md_t|$. 
The output of the generator is an alteration of the input ontology for which at least the given number of minimum cardinality diagnoses with the required cardinality exist. Furthermore, to introduce inconsistencies (incoherencies), the generator applies fault patterns randomly to the input ontology depending on their probabilities. 

In this experiment we took five fault patterns from a case study reported by Rector et al.~\cite{Rector2004} and assigned fault probabilities according to their observations of typical user errors. Thus we assumed that in cases (a) and (b) (see Section~\ref{sect:discrimination}), where an axiom includes some roles (i.e.\ property assertions), axiom descriptions are faulty with a probability of $0.025$, in cases (c) and (d) $0.01$ and in case (e) $0.001$. In each iteration, the generator randomly selected an axiom to be altered and applied a fault pattern. Following this, another axiom was selected using the concept taxonomy and altered correspondingly to introduce an inconsistency (incoherency). The fault patterns were randomly selected in each step using the probabilities provided above.

For instance, given the description of a randomly selected concept $A$ and the fault pattern ``misuse of negation'', we added the construct $\sqcap \neg X$ to the description of $A$, where $X$ is a new concept name. Next, we randomly selected concepts $B$ and $S$ such that $S\sqsubseteq A$ and $S \sqsubseteq B$ and added $\sqcap X$ to the description of $B$.
During the generation process, we applied the \textsc{HS-Tree} algorithm after each introduction of an incoherency/inconsistency to control two parameters: the minimum number of minimal cardinality diagnoses in the ontology and their cardinality. The generator continues to introduce incoherences/inconsistencies until the specified parameter values are reached. For instance, if the minimum number of minimum cardinality diagnoses is equal to $m=6$ and their cardinality is $|\md_t|=4$, then the generated ontology will include at least $6$ diagnoses of cardinality $4$ and possibly some additional number of minimal diagnoses of higher cardinalities.

The resulting faulty ontology as well as the fault patterns and their probabilities were inputs for the ontology debugger.  The acceptance threshold $\sigma$ was set to $0.95$ and the number of most probable minimal diagnoses $n$ was set to $9$.
In addition, one of the minimal diagnoses with the required cardinality was randomly selected as the target diagnosis. 
Note, the target ontology is not equal to the original ontology, but rather a corrected version of the altered one in which the faulty axioms were repaired by replacing them with their original (correct) versions according to the target diagnosis.
The tests were performed using the ontologies bike2 to bike9, bcs3, galen and galen2 from Racer's benchmark suite\footnote{Available at \url{http://www.racer-systems.com/products/download/benchmark.phtml}}.

The average results of the evaluation performed on each test ontology (presented in Figure~\ref{fig:results}) show that the entropy-based approach outperforms the ``split-in-half'' heuristic  as well as the random query selection strategy by more than 50\% for the $|\md_t|=2$ case due to its ability to estimate the probabilities of diagnoses and to stop once the target diagnosis crossed the acceptance threshold. On average the algorithm required $8$ seconds to generate a query. In addition, Figure~\ref{fig:results} shows that the number of queries required increases as the cardinality of the target diagnosis increases, regardless of the method. Despite this, the entropy-based approach remains better than the ``split-in-half'' method for diagnoses with increasing cardinality. The approach did however require more queries to discriminate between high cardinality diagnoses because in such cases more minimal conflicts were generated. Consequently, the debugger should consider more minimal diagnoses in order to identify the target one.

For the next test we selected seven real-world ontologies described in Tables~\ref{tab:motivation} and~\ref{tab:stats}\footnote{All experiments were performed on a PC with Core2 Duo (E8400), 3 Ghz with 8 Gb RAM, running Windows 7 and Java 6.}. Performance of both the entropy-based and ``split-in-half'' selection strategies was evaluated using a variety of different prior fault probabilities to investigate under which conditions the entropy-based method should be preferred.

\begin{figure}[tb]
 \centering
 	\includegraphics[width=\linewidth]{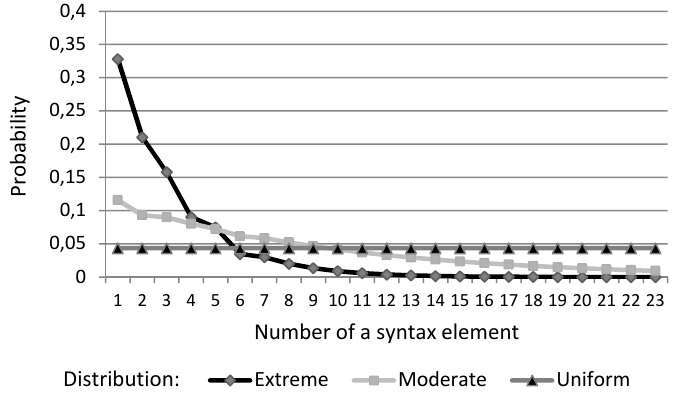}
 \caption{Example of prior fault probabilities of syntax elements sampled from extreme, moderate and uniform distributions.}
 \label{fig:priors}
\end{figure}

In our experiments we distinguished between three different distributions of prior fault probabilities: extreme, moderate and uniform (see Figure~\ref{fig:priors} for an example). The \emph{extreme distribution} simulates a situation in which very high failure probabilities are assigned to a small number of syntax elements. That is, the provider of the estimates is quite sure that exactly these elements are causing a fault. For instance, it may be well known that a user has problems formulating restrictions in OWL whereas all other elements, such as subsumption and conjunction, are well understood. In the case of a \emph{moderate distribution} the estimates provide a slight bias towards some syntax elements. This distribution has the same motivation as the extreme one, however, in this case the probability estimator is less sure about the sources of possible errors in axioms. Both extreme and moderate distributions correspond to the exponential distribution with $\lambda=1.75$ and $\lambda=0.5$ respectively. The \emph{uniform distribution} models the situation where no prior fault probabilities are provided and the system assigns equal probabilities to all syntax elements found in a faulty ontology. Of course the prior probabilities of diagnoses may not reflect the actual situation. Therefore, for each of the three distributions we differentiate between good, average and bad cases. In the \emph{good case} the estimates of the prior fault probabilities are correct and the target diagnosis is assigned a high probability. The \emph{average case} corresponds to the situation when the target diagnosis is neither favored nor penalized by the priors. In the \emph{bad case} the prior distribution is unreasonable and disfavors the target diagnosis by assigning it a low probability. 

We executed 30 tests for each of the combinations of the distributions and cases with an acceptance threshold $\sigma=0.85$ and a required number of most probable minimal diagnoses $n=9$. Each iteration started with the generation of a set of prior fault probabilities of syntax elements by sampling from a selected distribution (extreme, moderate or uniform). Given the priors we computed the set of all minimal diagnoses $\mD$ of a given ontology and selected the target one according to the chosen case (good, average or bad). In the good case the prior probabilities favor the target diagnosis and, therefore, it should be selected from the diagnoses with high probability. The set of diagnoses was ordered according to their probabilities and the algorithm iterated through the set starting from the most probable element. 
In the first iteration the most probable minimal diagnosis $\md_1$ is added to the set $G$. In next iteration $j$ a diagnosis $\md_j$ was added to the set $G$ if $\sum_{i \leq j}{p(\md_i)}\leq \frac{1}{3}$ and to the set $A$ if $\sum_{i\leq j}{p(\md_i)} \leq \frac{2}{3}$. The obtained set $G$ contained all most probable diagnoses which we considered as good. All diagnoses in the set $A\setminus G$ were classified as average and the remaining diagnoses $\mD \setminus A$ as bad. Depending on the selected case we randomly selected one of the diagnoses as the target from the appropriate set. 

The results of the evaluation presented in Table~\ref{tab:results} show that the entropy-based query selection approach clearly outperforms ``split-in-half'' in good and average cases for the three probability distributions. The average time required by the debugger to perform such basic operations as consistency checking, computation of minimal conflicts and diagnoses is presented in Table~\ref{tab:diagstats}. The results indicate that on average at most 17 seconds required to compute up to 9 minimal diagnoses and a query. Moreover, the number of axioms in a query remains reasonable in most of the cases stays bounds, i.e. between 1 and 4 axioms per query.

\begin{table*}[p] 
\centering
\begin{tabular}{l|c||ccc|ccc|ccc} 
\multicolumn{11}{c}{Entropy-based query selection} \\ \hline 
\multicolumn{1}{c}{Ontology}  & Case & \multicolumn{9}{c}{Distribution} \\ \cline{4-10}
\multicolumn{1}{c}{} & & \multicolumn{3}{c}{Extreme} 	 	 & \multicolumn{3}{c}{Moderate} 	 	 & \multicolumn{3}{c}{Uniform} \\ 	 	 
\multicolumn{1}{c}{}& & min & avg & \multicolumn{1}{c}{max} & min & avg & \multicolumn{1}{c}{max} & min & avg & \multicolumn{1}{c}{max} \\ \hline\hline
 & Good & 1 & \textbf{ 1.63 } & 3 & 1 & \textbf{ 1.7 } & 2 & 1 & \textbf{ 1.83 } & 2 \\ 
Chemical & Avg. & 1 & \textbf{ 1.87 } & 4 & 1 & \textbf{ 1.73 } & 3 & 1 & \textbf{ 1.7 } & 2 \\ 
 & Bad & 2 & 	3.03 	& 4 & 2 & 	3.03 	& 4 & 2 & 	3.17 	& 4 \\ \hline
 	 	 	 	 	 	 	 	 	 	 	 	 	 	
 & Good & 1 & \textbf{ 1.7 } & 3 & 1 & \textbf{ 2.4 } & 4 & 1 & \textbf{ 2.67 } & 3 \\ 
Koala & Avg. & 1 & \textbf{ 1.8 } & 3 & 1 & \textbf{ 2.37 } & 4 & 1 & \textbf{ 2.4 } & 3 \\ 
 & Bad & 1 & 	3.5 	& 6 & 2 & 	4.33 	& 7 & 3 & 	4.13 	& 5 \\ \hline
 	 	 	 	 	 	 	 	 	 	 	 	 	 	
 & Good & 1 & \textbf{ 3.27 } & 7 & 2 & \textbf{ 3.43 } & 7 & 3 & 	\textbf{ 3.87 } 	& 7 \\ 
Sweet-JPL & Avg. & 1 & \textbf{ 3.5 } & 6 & 1 & 	4.03 	& 7 & 3 & 	4.07 	& 6 \\ 
 & Bad & 3 & 	3.93 	& 6 & 2 & 	4.03 	& 6 & 3 & \textbf{ 3.37 } & 4 \\ \hline
 	 	 	 	 	 	 	 	 	 	 	 	 	 	
 & Good & 1 & \textbf{ 2.37 } & 4 & 2 & \textbf{ 2.73 } & 4 & 2 & \textbf{ 2.77 } & 3 \\ 
miniTambis & Avg. & 1 & \textbf{ 2.53 } & 4 & 2 & \textbf{ 4.03 } & 8 & 3 & \textbf{ 4.53 } & 7 \\ 
 & Bad & 3 & 	6.43 	& 11 & 3 & 	7.93 	& 17 & 5 & 	9.03 	& 13 \\ \hline
 	 	 	 	 	 	 	 	 	 	 	 	 	 	
 & Good & 1 & \textbf{ 2.7 } & 4 & 3 & \textbf{ 3.83 } & 7 & 3 & \textbf{ 4.4 } & 8 \\ 
University & Avg. & 1 & \textbf{ 3.4 } & 6 & 3 & 	7.03 	& 12 & 4 & \textbf{ 7.27 } & 10 \\ 
 & Bad & 5 & 	9.13 	& 15 & 5 & 	9.7 	& 14 & 6 & 	10.03 	& 14 \\ \hline
 	 	 	 	 	 	 	 	 	 	 	 	 	 	
 & Good & 1 & \textbf{ 3.2 } & 11 & 3 & \textbf{ 3.1 } & 4 & 3 & \textbf{ 3.93 } & 6 \\ 
Economy & Avg. & 1 & \textbf{ 4.63 } & 14 & 3 & \textbf{ 5.57 } & 12 & 5 & \textbf{ 6.5 } & 8 \\ 
 & Bad & 8 & 	12.3 	& 19 & 6 & 	11.5 	& 21 & 7 & 	11.67 	& 19 \\ \hline
 	 	 	 	 	 	 	 	 	 	 	 	 	 	
 & Good & 1 & \textbf{ 5.63 } & 14 & 1 & \textbf{ 6.97 } & 12 & 3 & \textbf{ 9.5 } & 14 \\ 
Transportation & Avg. & 1 & \textbf{ 6.9 } & 16 & 1 & \textbf{ 7.73 } & 12 & 3 & \textbf{ 8.73 } & 14 \\ 
 & Bad & 3 & \textbf{ 12.4 } & 18 & 8 & \textbf{ 12.8 } & 20 & 3 & \textbf{ 12.1 } & 18 \\ \hline

    \multicolumn{11}{c}{ } \\
 \multicolumn{11}{c}{ ``Split-in-half'' query selection} \\\hline\hline
 & Good & 2 & 	2.63 	& 3 & 2 & 	2.7 	& 3 & 2 & 	2.53 	& 3 \\ 
Chemical & Avg. & 2 & 	2.63 	& 3 & 2 & 	2.67 	& 3 & 2 & 	2.77 	& 3 \\ 
 & Bad & 2 & \textbf{ 2.63 } & 3 & 2 & \textbf{ 2.6 } & 3 & 2 & \textbf{ 2.4 } & 3 \\ \hline
 	 	 	 	 	 	 	 	 	 	 	 	 	 	
 & Good & 3 & 	3.3 	& 4 & 3 & 	3.3 	& 4 & 3 & 	3.47 	& 4 \\ 
Koala & Avg. & 3 & 	3.33 	& 4 & 3 & 	3.2 	& 4 & 3 & 	3.23 	& 4 \\ 
 & Bad & 3 & \textbf{ 3.43 } & 4 & 3 & \textbf{ 3.4 } & 4 & 3 & \textbf{ 3.5 } & 4 \\ \hline
 	 	 	 	 	 	 	 	 	 	 	 	 	 	
 & Good & 3 & 	3.83 	& 4 & 3 & 	3.8 	& 4 & 4 &  4  & 4 \\ 
Sweet-JPL & Avg. & 3 & 	3.57 	& 4 & 3 & \textbf{ 3.8 } & 4 & 3 & \textbf{ 3.47 } & 4 \\ 
 & Bad & 3 & \textbf{ 3.87 } & 4 & 3 & \textbf{ 3.8 } & 4 & 3 & 	3.8 	& 4 \\ \hline
 	 	 	 	 	 	 	 	 	 	 	 	 	 	
 & Good & 4 & 	5.33 	& 6 & 4 & 	5 	& 6 & 4 & 	4 	& 4 \\ 
miniTambis & Avg. & 4 & 	5.1 	& 6 & 4 & 	4.93 	& 7 & 5 & 	5.43 	& 7 \\ 
 & Bad & 5 & \textbf{ 5.93 } & 8 & 4 & \textbf{ 5.8 } & 7 & 5 & \textbf{ 6.3 } & 7 \\ \hline
 	 	 	 	 	 	 	 	 	 	 	 	 	 	
 & Good & 4 & 	5.93 	& 8 & 4 & 	6 	& 8 & 4 & 	5.43 	& 8 \\ 
University & Avg. & 4 & 	5.87 	& 7 & 5 & \textbf{ 6.73 } & 9 & 6 & 	7.37 	& 8 \\ 
 & Bad & 5 & \textbf{ 6.97 } & 9 & 5 & \textbf{ 7.2 } & 9 & 5 & \textbf{ 7 } & 8 \\ \hline
 	 	 	 	 	 	 	 	 	 	 	 	 	 	
 & Good & 6 & 	7.87 	& 11 & 6 & 	7.4 	& 10 & 6 & 	7.5 	& 10 \\ 
Economy & Avg. & 6 & 	8 	& 12 & 5 & 	7.63 	& 12 & 6 & 	8.73 	& 13 \\ 
 & Bad & 9 & \textbf{ 11.50 } & 14 & 6 & \textbf{ 11.1 } & 14 & 8 & \textbf{ 11.3 } & 15 \\ \hline
 	 	 	 	 	 	 	 	 	 	 	 	 	 	
 & Good & 5 & 	8.03 	& 13 & 5 & 	7.3 	& 11 & 6 & 	11.43 	& 18 \\ 
Transportation & Avg. & 3 & 	9 	& 16 & 5 & 	9.4 	& 13 & 5 & 	11.43 	& 18 \\ 
 & Bad & 10 & 	12.67 	& 19 & 7 & 	13 	& 19 & 6 & 	13.8 	& 20 \\ \hline

\end{tabular}
\caption{Minimum, average and maximum number of queries required by the entropy-based and ``split-in-half'' query selection methods to identify the target diagnosis in real-world ontologies. Ontologies are ordered by the number of diagnoses.}
\label{tab:results}
\end{table*}

\begin{table*}
	\centering
		\begin{tabular}{l||ccc|ccc|ccc}
  \multicolumn{1}{c}{Ontology}	&	\multicolumn{3}{c}{Good} &	\multicolumn{3}{c}{Average}&\multicolumn{3}{c}{Bad} \\ \cline{2-10}
  \multicolumn{1}{c}{}		&	DT	&	QT	&	QL	&	DT	&	QT	&	QL	&	DT	&	QT	&	QL	\\	\hline
Chemical	&	459.33	&	117.67	&	3	&	461.33	&	121	&	3.34	&	256.67	&	75.67	&	2.19	\\	
Koala 	&	88.33	&	1308.33	&	3.47	&	92	&	1568.67	&	3.90	&	56.33	&	869.33	&	2.36	\\	
Sweet-JPL	&	2387.33	&	691.67	&	1.48	&	2272	&	926	&	1.61	&	2103	&	1240.33	&	1.57	\\	
miniTabmis	&	481.33	&	2764.33	&	3.27	&	398.33	&	2892	&	2.53	&	238.67	&	3223	&	1.76	\\	
University	&	189.33	&	822.67	&	3.91	&	145	&	903.33	&	2.82	&	113	&	872	&	2.11	\\	
Economy	&	2953.33	&	6927	&	3.06	&	3239	&	8789	&	3.80	&	3083	&	8424.67	&	1.58	\\	
Transportation	&	6577.33	&	9426.33	&	2.37	&	7080.67	&	10135.33	&	2.29	&	7186.67	&	9599.67	&	1.64	\\	\hline
 
		\end{tabular}
	\caption{Average time required to compute at most nine minimal diagnoses (DT) and a query (QT) in each iteration, as well as the average number of axioms in a query after minimization (QL). The averages are shown for extreme, moderate and uniform distributions using the entropy-based query selection method. Time is measured in milliseconds.}
	\label{tab:diagstats}
\end{table*}

In the uniform case better results were observed since the diagnoses have different cardinality and structure, i.e. they include different syntax elements. Consequently, even if equal probabilities for all syntax elements (uniform distribution) are given, the probabilities of diagnoses are different. Axioms with a greater number of syntax elements receive a higher fault probability. Also, diagnoses with a smaller cardinality in many cases receive a higher probability. This information provides enough bias to favor the entropy-based method.

In the bad case, where the target diagnosis received a low probability and no information regarding the prior fault probabilities was given, we observed that the performance of the entropy-method improved as more queries were posed. In particular, in the University ontology the performance is essentially similar (7.27 vs.\ 7.37) whereas in the Economy and Transportation ontology the entropy-based method can save and average of two queries. 

``Split-in-half'' appears to be particularly inefficient in all good, average and bad cases when applied to ontologies with a large number of minimal diagnoses, such as Economy and Transportation. The main problem is that no stop criteria can be used with the greedy method as it is unable to provide any ordering on the set of diagnoses. Instead, the method continues until no further queries can be generated, i.e.\ only one minimal diagnosis exists or there are no discriminating queries. Conversely, the entropy-based method is able to improve its probability estimates using Bayes-updates as more queries are answered and to exploit the differences in the probabilities in order to decide when to stop. 

The most significant gains are achieved for ontologies with many minimal diagnoses and for the average and good cases, e.g. the target diagnosis is within the first or second third of the minimal diagnoses ranked by their prior probability. In these cases the entropy-based method can save up to 60\% of the queries. 

Therefore, we can conclude that even rough estimates of the prior fault probabilities are sufficient, provided that the target diagnosis is not significantly penalized. Even if no fault probabilities are available and there are many minimal diagnoses, the entropy-based method is advantageous.  The differences between probabilities of individual syntax elements appears not to influence the results of the query selection process and affect only the number of outliers, i.e.\ cases in which the diagnosis approach required either few or many queries compared to the average. 

Another interesting observation is that often both methods eliminated more than $n$ diagnoses in one iteration. For instance, in the case of the Transportation ontology both methods were able to remove hundreds of minimal diagnoses with a small number of queries. This behavior appears to stem from relations between the diagnoses. That is, the addition of a query to either $\Te$ or $\Tne$ allows the method to remove not only the diagnoses in sets $\dx$ or $\dnx$, but also some unobserved diagnoses that were not in any of the sets of $n$ leading diagnoses computed by \textsc{HS-Tree}. Given the sets $\Te$ and $\Tne$, \textsc{HS-Tree} automatically invalidates all diagnoses which do not fulfill the requirements (see Definition~\ref{def:diag}). 

\begin{figure}[tb]
 \centering
 	\includegraphics[width=\linewidth]{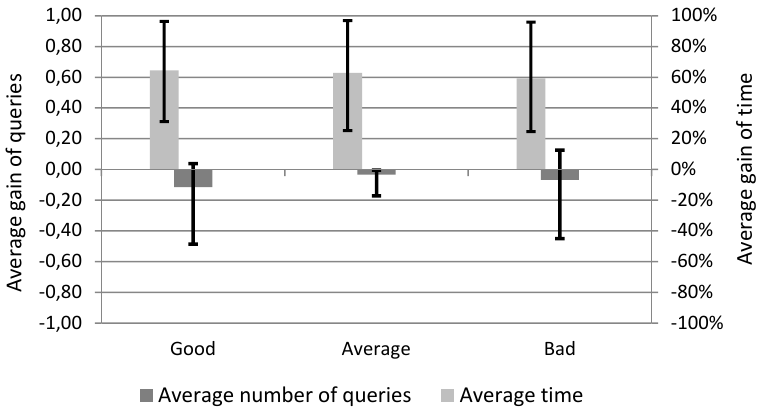}
 \caption{Average time/query gain resulting from the application of the extended CKK partitioning algorithm. The whiskers indicate the maximum and minimum possible average gain of queries/time using extended CKK.}
 \label{fig:greedy}
\end{figure}

\begin{figure}[tb]
 \centering
 	\includegraphics[width=.9\linewidth]{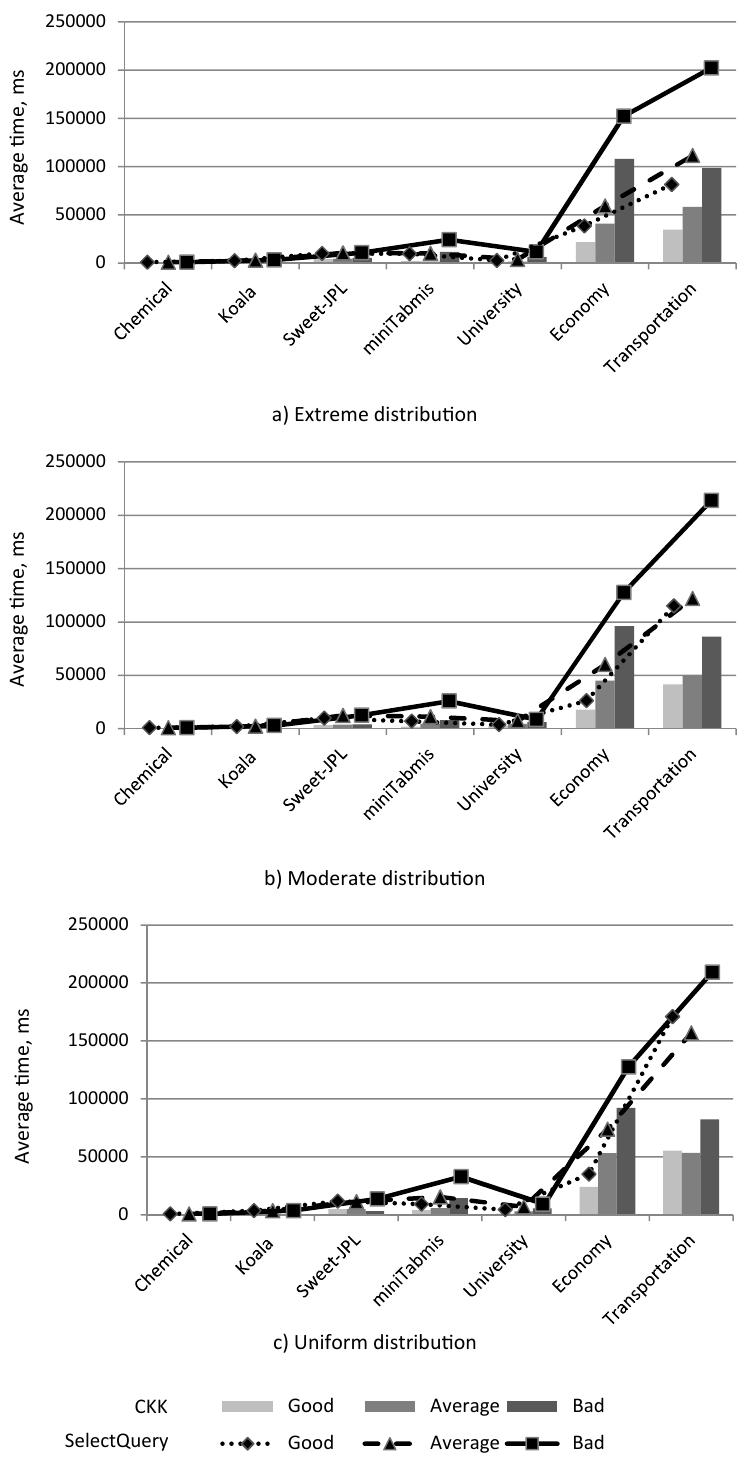}
 \caption{Average time required to identify the target diagnosis using CKK and brute force query selection algorithms.}
 \label{fig:averages}
\end{figure}

The extended CKK method presented in Section~\ref{sect:impl} was evaluated in the same settings as the complete Algorithm~\ref{algo:computeQuery} with acceptance threshold $\gamma=0.1$.  The obtained results presented in Figure~\ref{fig:greedy} show that the extended CKK method decreases the length of a debugging session by at least 60\% while requiring on average $0.1$ queries more than Algorithm~\ref{algo:computeQuery}. In some cases (mostly for the uniform distribution) the debugger using CKK search required even fewer queries than Algorithm~\ref{algo:computeQuery} because of the inherent uncertainty of the domain. The plot of the average time required by Algorithm~\ref{algo:computeQuery} and CKK to identify the target diagnosis presented in Figure~\ref{fig:averages} shows that the application of the latter can reduce runtime significantly.

In the last experiment we tried to simulate an expert developing large real-world ontologies\footnote{The ontologies taken from TONES repository {\tt http://owl.cs.manchester.ac.uk/repository}} as described in Table \ref{tab:bigstats}. Often in such settings an expert makes small changes to the ontology and then runs the reasoner to verify that the changes are valid, i.e. the ontology is consistent and its entailments are correct. To simulate this scenario we used the generator described in the first experiment to introduce 1 to 3 random changes that would make the ontology incoherent. Then, for each modified ontology, we performed 15 tests using the fault distributions as in the second test. The results obtained by the entropy-based query selection method using CKK for query computation are presented in Table~\ref{tab:bigres}. These results show that the method can be used for analysis of large ontologies with over 33000 axioms while requiring a user to wait for only a minute to compute the next query.

\begin{table*}[bt]
	\centering
		\begin{tabular} {l|cc} 
  \multicolumn{1}{c}{Ontology}	&	Cton	&	Opengalen-no-propchains	\\ \hline	
Axioms & 33203 & 9664  \\
DL	&	$\mathcal{SHF}$	&	$\mathcal{ALEHIF}^{(D)}$ \\	

\#CS/min/max  	&	6/3/7	&	9/5/8	\\	
\#D/min/max	&	15/1/5	&	110/2/6	\\	
Consistency	&	5/209/1078	&	1/98/471	\\	
QuickXplain	&	17565/20312/38594	&	7634/10175/12622	\\	
Diagnosis	&	1/5285/38594	&	10/1043/19543	\\	
Overall runtime	&	146186	&	119973	\\	\hline
   
		\end{tabular}
	\caption{Statistics for the real-world ontologies used in the stress-tests measured for a single random alteration. \#CS/min/max are the number of minimal conflict sets, and their minimum and maximum cardinality. The same notation is used for diagnoses \#D/min/max.  The minimum/average/maximum time required to make a consistency check (Consistency), compute a minimal conflict set (QuickXplain) and a minimal diagnosis are measured in milliseconds. Overall runtime indicates the time required to compute all minimal diagnoses in milliseconds.}
	\label{tab:bigstats}
\end{table*}

\begin{table*}[bt]
	\centering
		\begin{tabular} {l|ccccc} 
	 \multicolumn{6}{c}{Good} \\ \hline
					
  \multicolumn{1}{c|}{Ontology}	&	\#Query	&	Overall	&	QT	&	DT	&	QL \\ \hline
Cton	&	3	&	176828	&	6918	&	52237	&	4 \\
Opengalen-no-propchains	&	8	&	154145	&	2349	&	22905	&	4 \\\hline
		\multicolumn{6}{c}{Average}		\\\hline						
Cton	&	4	&	177383	&	6583	&	52586	&	3 \\
Opengalen-no-propchains	&	7	&	151048	&	3752	&	21344	&	4 \\\hline
		\multicolumn{6}{c}{Bad} 								\\\hline
Cton	&	5	&	190407	&	5742	&	35608	&	1 \\
Opengalen-no-propchains	&	14	&	177728	&	1991	&	11319	&	3 \\\hline

\end{tabular}
	\caption{Average values measured for extreme, moderate and uniform distributions in each of the good, average and bad cases. \#Query is the number of queries required to find the target diagnosis. Overall runtime as well as the time required to compute a query (QT) and at least nine minimal diagnoses (DT) are given in milliseconds. Query length (QL) shows the average number of axioms in a query.}
	\label{tab:bigres}
\end{table*}

\section{Related work}\label{sect:relwork}
Despite the range of ontology diagnosis methods available (see ~\cite{schlobach2007,Kalyanpur.Just.ISWC07,friedrich2005gdm}), to the best of our knowledge no interactive ontology debugging methods, such as our ``split-in-half" or entropy-based methods, have been proposed so far. The idea of ranking of diagnoses and proposing a target diagnosis is presented in~\cite{Kalyanpur2006}. This method uses a number of measures such as: (a) the frequency with which an axiom appears in conflict sets, (b) impact on an ontology in terms of its ``lost'' entailments when an axiom is modified or removed, (c) ranking of test cases, (d) provenance information about axioms, and (e) syntactic relevance. For each axiom in a conflict set, these measures are evaluated and combined to produce a rank value. These ranks are then used by a modified \textsc{HS-Tree} algorithm to identify diagnoses with a minimal rank. 
However, the method fails when a target diagnosis cannot be determined reliably with the given a-priori knowledge. In our work required information is acquired until the target diagnosis can be identified with confidence.
In general, the work of~\cite{Kalyanpur2006} can be combined with the ideas presented in this paper as axiom ranks can be taken into account together with other observations for calculating the prior probabilities of the diagnoses.

The idea of selecting the next best query based on the expected entropy was exploited in the generation of decisions trees in~\cite{Quinlan1986} and further refined for selecting measurements in the model-based diagnosis of circuits in~\cite{dekleer1987}. We extend these methods to  query selection in the domain of ontology debugging.  

In the area of debugging logic programs, Shapiro~\cite{Shapiro83} developed debugging methods based on query answering. Roughly speaking, Shapiro's method aims to detect one fault at a time by querying an oracle about the intended behavior of a Prolog program at hand. In our terminology, for each answer that must not be entailed this diagnosis approach generates one conflict at a time by exploiting the proof tree of a Prolog program. The method then identifies a query that splits the conflict in half.  Our approach can deal with multiple diagnoses and conflicts simultaneously which can be exploited by query generation strategies such as ``split-in-half" and entropy-based methods. Whereas the ``split-in-half" strategy splits the set of diagnoses in half, Shapiros's method focuses on one conflict. Furthermore, the exploitation of failure probabilities is not considered in~\cite{Shapiro83}. However, Shapiro's method includes the learning of new clauses in order to cover not entailed answers. Interleaving discrimination of diagnoses and learning of descriptions is currently not considered in our approach because of their additional computational costs.

From a general point of view Shapiro's method can be seen as a prominent example of inductive logic programming (ILP) including systems such as~\cite{MuggletonB88,Muggleton95}. In particular, \cite{Muggleton95} proposes inverse entailments combined with general to specific search through a refinement graph with the goal of generating a theory (hypothesis) which covers the examples and fulfills additional properties. 
Compared to ILP, the focus of our work lies on the theory revision. However, our knowledge representation languages are variants of description logics and not logic programs. 
%
%
Moreover, our method aims to discover axioms which must be changed while minimizing user interaction. Preferences  of theory changes are expressed by probabilities which are updated through Bayes' rule. Other preferences based on plausible extensions of the theory were not considered, again because of their computational costs. 

Although model-based diagnosis has also been applied to logic programs~\cite{ConsoleFD93}, constraint knowledge-bases~\cite{FelfernigFJS04} and hardware descriptions~\cite{FriedrichSW99}, none of these approaches propose a query generation method to discriminate between diagnoses.

\section{Conclusions}

In this paper we presented an approach to the interactive diagnosis of ontologies. This approach is applicable to any ontology language with monotonic semantics.
We showed that the axioms generated by classification and realization reasoning services can be exploited to generate queries which differentiate between diagnoses. For selecting the best next query we proposed two strategies: The ``split-in-half" strategy prefers queries which allow eliminating a half of leading diagnoses. 
The entropy-based strategy employs information theoretic concepts to exploit knowledge about the likelihood of axioms needing to be changed because the ontology at hand is faulty. 
Based on the probability of an axiom containing an error we predict the information gain produced by a query result, enabling us to select the best subsequent query according to a one-step-lookahead entropy-based scoring function. We described the implementation of a interactive debugging algorithm and compared the entropy-based method with the ``split-in-half" strategy. Our experiments showed a significant reduction in the number of queries required to identify the target diagnosis when the entropy-based method is applied. Depending on the quality of the prior probabilities the number of queries required may be reduced by up to 60\%. 

In order to evaluate the robustness of the entropy-based method we experimented with different prior fault probability distributions as well as different qualities of the prior probabilities. Furthermore, we investigated cases where knowledge about failure probabilities is missing or inaccurate. Where such knowledge is unavailable, the entropy-based methods ranks the diagnoses based on the number of syntax elements contained in an axiom and the number of axioms in a diagnosis. If we assume that this is a reasonable guess (i.e.\ the target diagnosis is not at the lower end of the diagnoses ranked by their prior probabilities) then the entropy-based method outperforms ``split-in-half". Moreover, even if the initial guess is not reasonable, the entropy-based method improves the accuracy of the probabilities as more questions are asked.
Furthermore, the applicability of the approach to real-world ontologies containing thousand of axioms was demonstrated by extensive set of evaluations which are publicly available.

\bibliographystyle{elsarticle-num}
\bibliography{jws11}

\end{document}